\documentclass[sigconf]{acmart}

\usepackage{booktabs}
\usepackage{graphicx}
\usepackage{color}

\usepackage{bm}

\AtBeginDocument{%
  }

\copyrightyear{2025}
\acmYear{2025}
\setcopyright{acmlicensed}
\acmConference[WSDM '25]{Proceedings of the
	Eighteenth ACM International Conference on Web Search and Data
	Mining}{March 10--14, 2025}{Hannover, Germany}
\acmBooktitle{Proceedings of the Eighteenth ACM International Conference on
	Web Search and Data Mining (WSDM '25), March 10--14, 2025, Hannover,
	Germany}
\acmDOI{10.1145/3701551.3703526}
\acmISBN{979-8-4007-1329-3/25/03}

\begin{document}

\title{ESA: Example Sieve Approach for Multi-Positive and Unlabeled Learning}

\author{Zhongnian Li}
\orcid{1234-5678-9012}
\affiliation{%
	  \institution{$^{\dag}$School of Computer Science and Technology, China University of Mining and Technology}
	  \city{Xuzhou}
	  \country{China}\\
	  \institution{$^{\ddag}$Mine Digitization Engineering Research Center of the Ministry of Education}
	  \city{Xuzhou}
	  \country{China}
	}
\email{zhongnianli@cumt.edu.cn}

\author{Meng Wei}
\orcid{0009-0000-3836-6487}
\affiliation{%
	\institution{$^{\dag}$School of Computer Science and Technology, China University of Mining and Technology}
	\city{Xuzhou}
	\country{China}\\
	\institution{$^{\ddag}$Mine Digitization Engineering Research Center of the Ministry of Education}
	\city{Xuzhou}
	\country{China}
}
\email{mengw@cumt.edu.cn}

\author{Peng Ying}
\orcid{0009-0005-8007-5583}
\affiliation{%
	\institution{$^{\dag}$School of Computer Science and Technology, China University of Mining and Technology}
	\city{Xuzhou}
	\country{China}\\
	\institution{$^{\ddag}$Mine Digitization Engineering Research Center of the Ministry of Education}
	\city{Xuzhou}
	\country{China}
}
\email{pengying@cumt.edu.cn}

\author{Xinzheng Xu}
\authornote{Corresponding author.}
\orcid{0000-0001-6973-799X}
\affiliation{%
	\institution{$^{\dag}$School of Computer Science and Technology, China University of Mining and Technology}
	\city{Xuzhou}
	\country{China}\\
	\institution{$^{\ddag}$State Key Lab. for Novel Software Technology, Nanjing University}
	\city{Nanjing}
	\country{China}\\
	\institution{$^{\S}$Mine Digitization Engineering Research Center of the Ministry of Education}
	\city{Xuzhou}
	\country{China}
}
\email{xxzheng@cumt.edu.cn}

\renewcommand{\shortauthors}{Zhongnian Li, Meng Wei, Peng Ying, $\&$ Xinzheng Xu}

\begin{abstract}
	Learning from Multi-Positive and Unlabeled (MPU) data  has gradually attracted significant attention from practical applications.  Unfortunately,  the risk of MPU also suffer from the shift of minimum risk,  particularly when the models are very flexible as shown in Fig.\ref{moti}.  In this paper, to alleviate the  shifting of minimum risk problem, we propose an Example Sieve Approach (ESA) to select examples for training a multi-class classifier. Specifically, we sieve out some examples by utilizing the Certain Loss (CL) value of each example  in the training stage and analyze the consistency of the proposed risk estimator. Besides, we show that the  estimation error of proposed ESA obtains the optimal parametric convergence rate. Extensive experiments on various real-world datasets show the proposed approach outperforms previous methods. Source code is available at  \href{https://github.com/WilsonMqz/ESA}{https://github.com/WilsonMqz/ESA}
\end{abstract}





\maketitle

\section{Introduction}
Learning from Positive and Unlabeled (PU) data \cite{DBLP:journals/ml/BekkerD20,DBLP:conf/pkdd/BekkerRD19,DBLP:conf/nips/PlessisNS14, DBLP:journals/ml/PlessisNS17, DBLP:journals/pami/GongSLZYT21,DBLP:conf/nips/WangC0023,DBLP:conf/nips/WangWGLC23,DBLP:conf/kdd/ZhuW0DZD0LR023, DBLP:conf/kdd/PeriniVD23} focuses on training a binary classifier  using only positive and unlabeled data. The PU \cite{DBLP:conf/icml/PlessisNS15,DBLP:journals/vldb/GalarragaTHS15,DBLP:conf/aaai/GongWYXL18, DBLP:conf/iclr/KatoTH19,DBLP:conf/iccv/ZhaoWLZ23,DBLP:conf/ecai/MielniczukW23,DBLP:conf/ecai/FurmanczykMRT23,DBLP:conf/cikm/YangZYK23, sevetlidis2024dense} problem arises in various practical applications, such as outlier detection \cite{DBLP:journals/ker/KhanM14,DBLP:conf/aaai/KhotNS14,DBLP:conf/aaai/LiangZ0WZYWHZ23,DBLP:conf/aaai/WangPZWH24} and has gradually attracted significant attention in the computer vision \cite{DBLP:conf/ijcai/LatulippeDGL13,DBLP:journals/tgrs/LiGE11,DBLP:conf/cikm/YangZYK23,DBLP:conf/aaai/DaiLZYL23} and  pattern recognition communities \cite{DBLP:journals/candc/JowkarM16,DBLP:journals/prl/MordeletV14,DBLP:conf/icdm/VercruyssenMVMB18,DBLP:conf/aaai/CaoR0SZ24}. Recently, learning from Multi-Positive and Unlabeled (MPU) \cite{DBLP:conf/icdm/ShuLY020,DBLP:conf/ijcai/XuX0T17, perini2023learning} is proposed to solve multi-class classification which is more common than binary classification in real-world applications.  Learning from MPU aims to train  multi-class classifiers using multi-positive classes and unlabeled data without negative data.

Existing MPU works \cite{DBLP:conf/icdm/ShuLY020,DBLP:conf/ijcai/XuX0T17, DBLP:conf/acl/ZhouLL22} solve the multi-class classification problem using a rewritten risk estimator, which is a powerful tool to evaluate the classification risk over training data. Xu \emph{et al.}\cite{DBLP:conf/ijcai/XuX0T17} propose  an unbiased risk estimator to learn from labeled and unlabeled data and provide generalization error bound. Unfortunately, the unbiased risk estimator of MPU \cite{DBLP:conf/icdm/ShuLY020} leads to overfitting, and even empirical risks go negative as the models become  more flexible.  Shu \emph{et al.}\cite{DBLP:conf/icdm/ShuLY020} propose an alternative estimator to avoid negative empirical risks by modifying the training loss. The alternative estimator partially alleviates overfitting in the training stage, thereby addressing the issue with the unbiased risk estimator.

\begin{figure}
	\centering
	\includegraphics[width=0.5\textwidth]{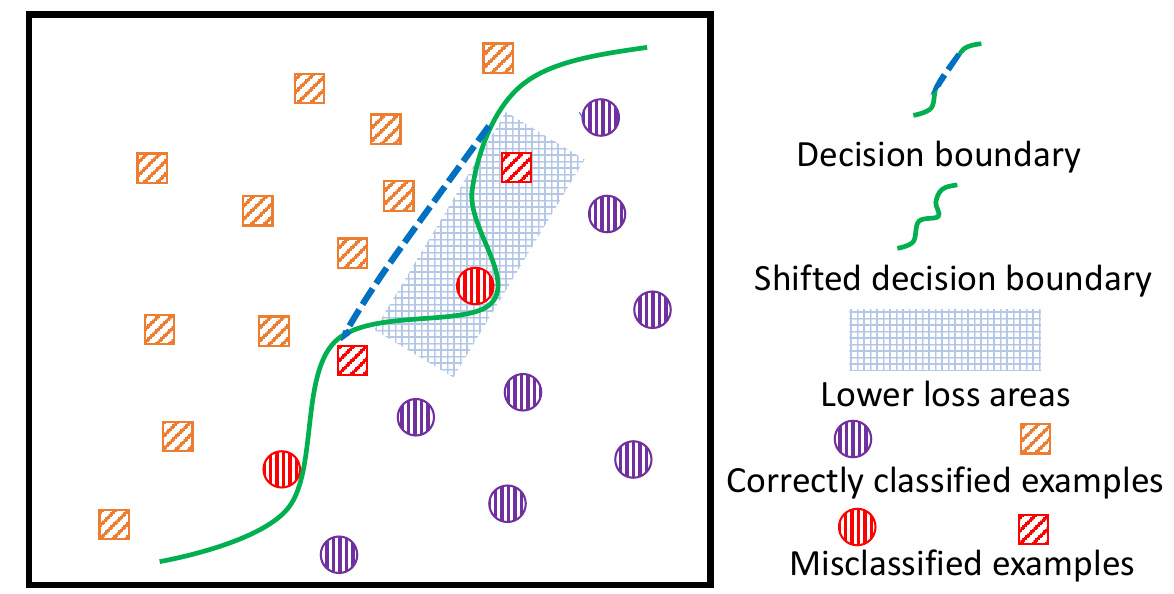}
	\caption{Illustrating the shift of minimum risk using unlabeled data, particularly when using complex models such as deep networks to train classifiers. Despite the misclassification of some examples, the decision boundary persists in traversing areas of lower loss, resulting in a shift of the minimum risk.  Furthermore, while an increase in the number of examples with lower loss results in a reduction of minimum risk, the shifting of minimum risk leads to overfitting in learning from multi-positive and unlabeled data. }
	\label{moti}
\end{figure}
In fact,  empirical risks also suffer from the problem of shifting of minimum risk (i.e., misclassifying examples near the decision boundary), as shown in Fig.\ref{moti}. Despite the misclassification of some examples, the decision boundary still tends to pass through areas with lower loss. This problem is hard to avoid in the training stage and may degrade the performance of the unbiased risk estimator.  Unfortunately, existing MPU methods have yet to explore the problem of the shifting of minimum risk, which inspires us to exploit examples risk to address this problem. 

To alleviate the severe  problem  of  minimum risk shifting, this paper investigates  how to select appropriate examples in the training stage to reduce the misclassification of some examples  near the decision boundary. Different from the above line of work, we propose an Example Sieve Approach (ESA) which  utilizes the loss values of each example to sieve out some examples from the training data.  Compared to the strategy of avoiding negative empirical risks to reduce overfitting, ESA is capable of addressing minimum risk shifting, thus enhancing the performance of the trained classifier. 

In this paper, we modify the unbiased risk estimator by utilizing sieved training data to alleviate poor empirical performance. By utilizing the Certain Loss (CL) of each example, we construct a biased empirical risk by sieving out certain overfitting examples with small loss values in the training data. We also prove that the biased risk estimator is consistent which means that minimizing empirical risk will obtain optimal multi-class classifiers. Furthermore, the analysis of the generalization error bound further justifies the capability of ESA in effectively utilizing labeled and unlabeled data. Moreover, extensive experiments conducted on various real-world datasets demonstrate that the proposed approach outperforms previous methods.
\section{Preliminaries}
\subsection{Learning from Multi-Positive and Unlabeled}
Supposing the data are collected from $C$ classes, the first $C-1$ classes are positive and the $C$-th class is regarded as negative. In the task of learning from multi-positive and unlabeled data, we consider that the labeled data are sampled from $C-1$ positive classes and the unlabeled data are collected from positive or negative classes.  Let $x\in \mathbb{R}^d$ and $y\in \{1,...,C\}$ be the input and output random variables following class-probability density $p(x,y)$.  The multi-positive dataset $D_m = \{(x_i^m,y_i)\}_{i=1}^{n_m}$ are sampled from the density  $p_m(x,y) = p(x,y \neq C)$, and unlabeled dataset $D_u = \{x_i^u\}_{i=1}^{n_u}$ are sampled from the distribution $p_u(x)=\sum\limits_{i = 1}^C {{\pi _i}p(x\left| {y = i)} \right.}$, where $\pi _i = p(y=i)$ denotes the class-prior probability. %defined over $\mathcal{X} \times \mathcal{Y}$. 

The goal of learning from multi-positive and unlabeled data \cite{DBLP:conf/icdm/ShuLY020,DBLP:conf/ijcai/XuX0T17, DBLP:conf/acl/ZhouLL22} is to train a classifier $\bm{f}:x \mapsto y$ to achieve good generalization ability for testing dataset. Xu \emph{et al.}\cite{DBLP:conf/ijcai/XuX0T17} propose an expected misclassification rate as follows:
\begin{equation}
	\begin{split}
		R(\bm{f}) = &\sum\limits_{i = 1}^{C - 1} {{\pi _i}p_i(\bm{f}(x) \ne i)}  + p_u(\bm{f}(x) \ne C) \\& - \sum\limits_{i = 1}^{C - 1} {{\pi _i}p_i(\bm{f}(x) \ne C)} 
	\end{split}
\end{equation}where $p_i(f(x)\ne i)$ denotes expected misclassification rate on the $i$-th class,  $p_u(f(x) \ne C)$ denotes the probability that unlabeled example has not been recognized as the $C$-th class and $p_i(f(x) \ne C)$ denotes the probability that $i$-th class example has not been classified as the $C$-th class. 

Then, shu \emph{et al.}\cite{DBLP:conf/icdm/ShuLY020} present an unbiased risk estimator to learn from multi-positive and unlabeled data as follows:
\begin{equation}\label{mpu}
	\begin{split}
		R_{mpu}(\bm{f}) = &\sum\limits_{i = 1}^{C - 1} {\pi _i}{\mathbb{E}_i}[\underbrace{L(\bm{f}(x),y = i)   - L(\bm{f}(x),y = C)}_{\text{Multi-positive Loss}}] \\& + {\mathbb{E}_{p_u}}[\underbrace{L(\bm{f}(x),y = C)}_{\text{Unlabeled Loss}}]
	\end{split}
\end{equation} where $\mathbb{E}_i$ denotes the abbreviation of $\mathbb{E}_{p(x|y=i)}$ and $\mathbb{E}_{p_u}$ denotes the abbreviation of $\mathbb{E}_{p_u(x)}$. In this paper, the proposed method uses the multi-class loss function $L(\bm{f}(x),y)$ to train the classifier. Then, accord to Eq.(\ref{mpu}), empirical approximation of unbiased risk estimator can be obtained as follows:
\begin{equation} \label{empu}
	\hat R_{mpu}(\bm{f}) = \sum\limits_{i = 1}^{C - 1} {\pi _i}\big(\hat R_i^i (\bm{f}) -\hat R_i^ C (\bm{f})\big) + \hat R_u^C (\bm{f}) 
\end{equation} where  $ \hat R_u^C (\bm{f}) = {1 \mathord{\left/
		{\vphantom {1 {{n_X}}}} \right.
		\kern-\nulldelimiterspace} {{n_u}}}\sum\nolimits_{i = 1}^{{n_u}} {L(f({x_i^u}),C)} $,  $\\ \hat R_i^C (\bm{f}) = {1 \mathord{\left/
		{\vphantom {1 {{n_i}}}} \right.
		\kern-\nulldelimiterspace} {{n_i}}}\sum\nolimits_{j = 1}^{{n_i}} {L(f({x_j^m}),C)} $ and  $ \hat R_i^i (\bm{f})={1 \mathord{\left/
		{\vphantom {1 {{n_i}}}} \right.
		\kern-\nulldelimiterspace} {{n_i}}}\sum\nolimits_{j = 1}^{{n_i}} {L(f({x_j^m}),i)}$ denotes empirical risk on data of the i-$th$ class. 

%Then, the one-versus-all strategy can be used for this method, i.e., $L(\bm{f}(x),y) = l(f_y(x))+ \frac{1}{{C - 1}}\sum\limits_{y' \ne y} {l( - {f_{y'}}(x))}$, which enjoys theoretical guarantees and  practical performance. The classification risk can be rewritten as
%\begin{equation}\label{ure}
%	\begin{split}
	%R(\bm{f}) =& \sum\limits_{i = 1}^{C - 1} {{\pi _i}{E_i}\bigg[l\big({f_i}(x)\big)}  - l\big({f_C}(x)\big) \\& + \frac{1}{{C - 1}}\big[l( - {f_C}(x)) - l( - {f_i}(x)\big]\bigg] \\& + {E_X}\big[l({f_C}(x)) + \frac{1}{{C - 1}}\sum\limits_{y \ne C} {l( - {f_y}(x))} \big]
	%	\end{split}
%\end{equation}
\subsection{Overfitting and Negative Risk}
Unbiased Risk Estimator (URE) is a powerful tool for training multi-class classifiers, which enables generalization error bounds to guarantee consistency. However, UREs suffer from severe overfitting during training and even lead to negative risk when the classifiers are complex models like deep networks \cite{DBLP:conf/icml/Chou0LS20,DBLP:conf/nips/KiryoNPS17,DBLP:conf/icdm/ShuLY020}. 

Kiryo \emph{et al.}\cite{DBLP:conf/nips/KiryoNPS17} propose a non-negative risk estimator for learning from positive and unlabeled data as follows:
\begin{equation}
	\hat R(\bm{f}) = {\pi _ + }\hat R_p^ + (\bm{f}) + \max (0,\; \hat R_u^ - (\bm{f}) - {\pi _ + }\hat R_p^ - (\bm{f}))
\end{equation} where $\hat R_p^ +(\bm{f})$ denotes the empirical risk of ${\mathbb{E}_+}[L(\bm{f}(x),y = +1)]$, $\hat R_p^ - (\bm{f})$ denotes the empirical risk of  ${\mathbb{E}_+}[L(\bm{f}(x),y = -1)]$ and $\hat R_u^ - (\bm{f})$ denotes the empirical risk of ${\mathbb{E}_{p_u}}[L(\bm{f}(x),y = -1)]$.  The risk estimator is biased yet optimal for training binary classifiers and its risk minimizer achieves  the same order of estimation error bound to unbiased counterparts.

Besides, Shu  \emph{et al.}\cite{DBLP:conf/icdm/ShuLY020} show that the classification risk is unbounded below,  which demonstrates that the unbiased empirical risk estimator suffers from  overfitting for learning a  classifier from multi-class positive and unlabeled data. To address this problem, They introduce  an alternative risk estimator which substitutes $-l(z)$ with $l(-z)$ to avoid the negative risk. Then, the classification risk obtains the lower bound. 
% \begin{equation}
	% 	\begin{split}
		% 	R(f) =& \sum\limits_{i = 1}^{C - 1} {{\pi _i}{E_i}[l({f_i}(x))}  + l( - {f_C}(x)) \\& + \frac{1}{{C - 1}}(l( - {f_C}(x)) + l({f_i}(x))]\\& + {E_X}[l({f_C}(x)) + \frac{1}{{C - 1}}\sum\limits_{y \ne C} {l( - {f_y}(x))} ]
		% \end{split}
	% \end{equation} 

In learning with complementary labels, Chou \emph{et al.}\cite{DBLP:conf/icml/Chou0LS20} focus on understanding how UREs lead to overfitting. They use an experiment to show how the complementary label distribution cause negative empirical risk for training classifiers and propose surrogate complementary loss to estimate the better gradients. 

While these methods address the overfitting issue by circumventing negative risk, they have not yet delved into the challenge of minimum risk shifting, which leads to misclassification of certain examples near the decision boundary.
\section{Example Sieve Approach} 
In this section, we propose an example sieve approach for learning with multi-positive and unlabeled data.
\subsection{Sieving Mechanisms}
\begin{figure}
		\centering
		\includegraphics[width=0.5\textwidth]{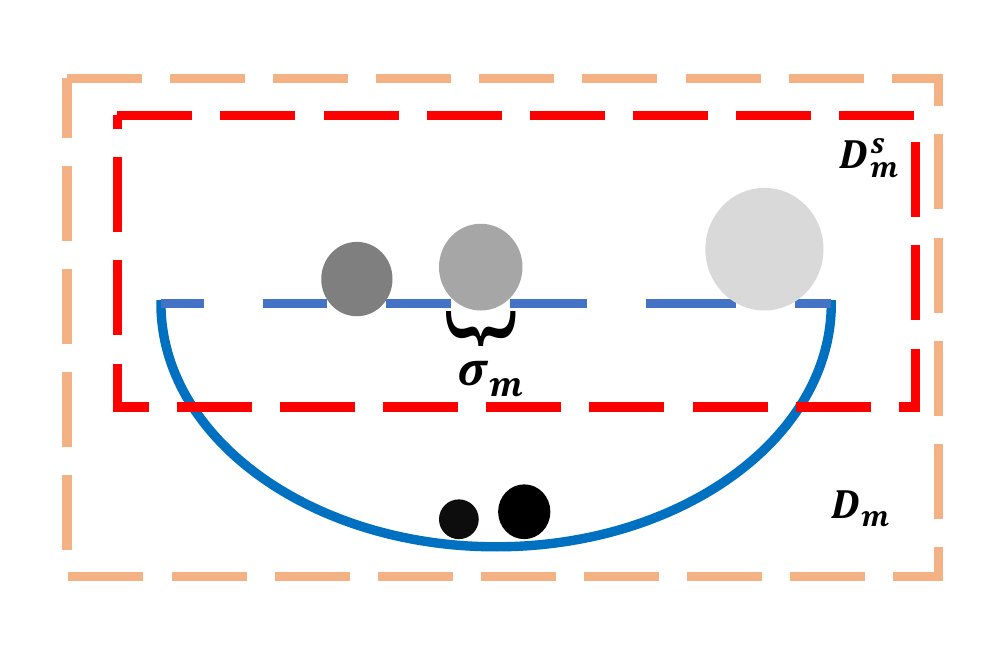}
		\caption{Illustrations of the generation procedure of sieved dataset $D^s_m$. Each circle denotes an example, and $\sigma_m$  denotes the lower bound. If the value of certain loss is smaller than lower bound,  the example is sieved out in the training stage.}
		\label{sieve}
	\end{figure}

To alleviate the severe problem  of  minimum risk shifting, we study how to sieve out appropriate examples for reducing overfitting.  In this paper, we use Certain Loss (CL) value of each example to sieve out some overfitting examples. Then, fix $\bm{f}$, according to Eq.(\ref{mpu}), the sieved multi-positived $  D_m^s $ and unlabeled dataset $ D_u^s $ can be given as follows: 
\begin{equation}\label{data^s}
	\begin{split}
		D^s_m = &\{ ({x_i},{y_i}) | CLm( \bm{f},(x_i,y_i))  \ge \sigma_m \}_{i=1}^{n_m^s} \subseteq D_m
	\end{split} 
\end{equation}
\begin{equation}\label{data_u}
	D^s_u = \{ {x_i} |  CLu(\bm{f},x_i) \ge {\sigma _u}\}_{i=1}^{n_u^s} \subseteq D_u
\end{equation} where 
\begin{equation}\label{cLm}
	\begin{split}
		CLm(\bm{f},(x_i,y_i))  = L(\bm{f}(x_i), y_i)   - L(\bm{f}(x_i), C)
	\end{split}
\end{equation}
\begin{equation}\label{cLu}
	\begin{split}
		CLu(\bm{f}, x_i) = L(\bm{f}(x_i),C)
	\end{split}
\end{equation} 
%\begin{equation}
%	\begin{split}
	%CLm(\bm{f},(x_i,y_i)) & = l\big({f_i}(x)\big)  - l\big({f_C}(x)\big)\\ & + \frac{1}{{C - 1}}\big[l( - {f_C}(x)) - l(-{f_i}(x))\big]
	%\end{split}
	%\end{equation}
	%\begin{equation}
	%	\begin{split}
		%	CLu(\bm{f},(x_i,y_i)) = l({f_C}({x_i})) + \frac{1}{{C - 1}}\sum\limits_{y \ne C} l( - {f_y}({x_i}) )
		%	\end{split}
	%\end{equation}  
	$\sigma_m$ and $\sigma_u $ denote the CL lower bounds, $ L(\bullet)$ denotes the multi-class loss function,  $n_m^s$ and $n_u^s$ denote the number of examples, and $\bm{f}$ denotes multi-class classifiers $(f_1,...,f_C)$. The generation procedure of sieved dataset $D^s_m$ is illustrated in Fig.\ref{sieve}.
	
	\noindent
	\textbf{ The distribution of  sieved dataset.} In contrast to class-probability density without sieving out some examples, the sieved dataset are draw i.i.d from two densities $p^s_m(x,y)$ and $p^s_u(x)$ as follows:
	\begin{equation}
		\begin{split}
			p^s_m(x,y) = 
			{{p_m(x,y  \left| CLm( \bm{f},(x,y))  \ge \sigma_m  \right.)} \bigg / { I^s_m }}
		\end{split} 
	\end{equation}
	\begin{equation}
		\begin{split}
			p^s_u(x)  = 
			{{p_u(x \left|  CLu( \bm{f}, x)  \ge \sigma_u  \right.)} 
				\bigg /} { I^s_u }
		\end{split} 
	\end{equation} where  $I^s_m = {\sum\limits_y {\int\limits_x {p_m(x,y  \left|  CLm( \bm{f},(x,y))  \ge \sigma_m  \right.)dx} } } $ and $\\ I^s_u = {\int\limits_x {p_u(x \left|  CLu( \bm{f}, x)  \ge \sigma_u  \right.)dx} } $ denote the distribution normalization.
	\subsection{Example sieve risk estimator}
	In this part, we develop an example sieve risk estimator for learning from multi-positive and unlabeled data. We first introduce an example sieved risk estimator. Then, we provide the details of the convex example sieve estimator. 
	
	\noindent
	\textbf{ Example sieve risk estimator.} Assuming that we  know sieved multi-positive and unlabeled data  densities $p^s_m(x,y)$ and $p^s_u(x)$,  we adopt the one-versus-rest (OVR) strategy\cite{DBLP:conf/nips/Zhang0MZ20} i.e.,
	\begin{equation}\label{ovr}
		L(f_1,...,f_C,x,y) = \phi(f_y(x))+\sum\limits_{y'=1,y' \ne y}^{y'=C} {\phi( - {f_{y'}}(x))}
	\end{equation} where $\phi: \mathbb R \mapsto [0, + \infty  )$ denotes a binary surrogate loss. Then, according to Eq.(\ref{mpu}), we can obtain the OVR risk for  sieving multi-class classification by rewriting an equivalent example sieve risk $R_{SEA}$ for multi-positive and unlabeled learning.
	
	\noindent
	\textbf{Proposition 1.} Let  $ p^s_i = p^s(x|y=i)$ denote a conditional density of sieved dataset, then the risk of example sieve can be equivalently represented as:
	\begin{equation}\label{sre}
		\begin{split}
			R_{SEA}&(f_1,...,f_C) = \sum\limits_{i = 1}^{C - 1} {{\pi _i}{\mathbb{E}_{p^s_i}}\bigg[\phi\big({f_i}(x)\big)}  - \phi\big({f_C}(x)\big) \\& + \big[\phi( - {f_C}(x)) - \phi( - {f_i}(x)\big]\bigg] \\& + {\mathbb{E}_{p^s_u}}\big[\phi({f_C}(x)) + \sum\limits_{y \ne C} {\phi( - {f_y}(x))} \big]
		\end{split}
	\end{equation} Proof can be found in Appendix. The OVR strategy obtains the prediction by $\bm f(x) = argmax_{f\in\{1,...,C\}}f_k(x)$. It is worth noting that  we can alleviate the overfitting  by selecting an appropriate lower bounds $\sigma_m$ and $\sigma_u $ in the above derivation. 
	
	\noindent
	\textbf{ Convex Example sieve risk estimator.} Due to the fact that $R_{SEA}$ is the non-convex caused by term $-\phi\big({f_C}(x)\big)$ and $- \phi( - {f_i}(x))$ in Eq.(\ref{sre})\cite{DBLP:conf/nips/Zhang0MZ20},  we now present a convex example sieve risk estimator w.r.t the binary classifiers when we use convex surrogate loss $\phi(\bullet)$. To eliminate the non-convexity, we select the surrogate loss carefully which satisfy $\phi(z)-\phi(-z) = -z$ for all $z \in \mathbb R$. Then the convex formulation of $R_{ESA}$ can be obtained as follows:
	\begin{equation}\label{csre}
		\begin{split}
			R_{SEA}&(f_1,...,f_C) = \sum\limits_{i = 1}^{C - 1} {{\pi _i}{\mathbb{E}_{p^s_i}}\big[{f_C}(x)}  - f_i(x)\big] \\& + {\mathbb{E}_{p^s_u}}\big[\phi({f_C}(x)) + \sum\limits_{y \ne C} {\phi( - {f_y}(x))} \big]
		\end{split}
	\end{equation}
	There are many surrogate losses satisfying the above condition, such as square loss $\phi(z) = (1-z)^2/4$ and  logistic loss $\phi(z) = log(1+exp(-z))$. We note that more multi-class and surrogate binary loss can be used when the convexity is not required for training classifiers.
	\subsection{Implementation}
	%	In this section, we introduce a practical algorithm for the proposed example sieve approach for learning with multi-positive and unlabeled data. Besides, we investigate implementation when using deep neural networks as a practical model.
	
	Given the sieved multi-positive and unlabeled dataset $D_m^s$ and $D_u^s$, we can obtain empirical approximation of the sieve risk estimator as follows:
	\begin{equation}\label{esre}
		\begin{split}
			\hat R_{SEA}&(f_1,...,f_C) = \sum\limits_{i = 1}^{C - 1} \frac{{{\pi _i}}}{{n_i^s}}\sum\limits_{{x_k} \in {D^s_i}} [\phi({f_i}({x_k})) \\& - \phi( {f_C}({x_k}))  + \phi( - {f_C}({x_k})) - \phi({f_i}({x_k})]  \\& + \frac{1}{{n_u^s}}\sum\limits_{{x_{k'}} \in {D^s_u}} {[\phi({f_C}({x_{k'}})) + \sum\limits_{y \ne C} {\phi( - {f_y}({x_{k'}}))} ]} 
		\end{split}
	\end{equation} where ${D^s_i}$ denotes the i-$th$ positive class dataset,  $n_i^s$  and $n_u^s$ denote the number of examples in dataset ${D^s_i}$ and $D^s_u$.
	
	We implement our approach by using deep neural networks.  Since the Eq.(\ref{esre}) can be optimized by the stochastic optimization method with a mini-batch, we use OVR strategy with the deep model and directly minimize the non-convex formulation of empirical risk. In this implementation, due to the  deep model, the final object function is non-convex, which inspire us to use more optimization methods to avoid the overfitting. Furthermore, many Mixture Proportion Estimation (MPE)\cite{DBLP:conf/icml/RamaswamyST16,DBLP:journals/ml/PlessisNS17} works can be used to estimate each class prior $\pi _i$ for Example sieve risk. 
	%	\begin{algorithm}[tb]
		%		\caption{ Example Sieve Approach }
		%		\label{al_esa}
		%		\begin{algorithmic}
			%			\STATE {\bfseries Input:}  { Multi-positive and unlabeled dataset $D$, the classifier $f_1,...,f_C$, the optimizer $O$ and CL lower bounds $\sigma_m$ and $\sigma_u $.}
			%			\STATE {\bfseries Output:} Model parameters $w$
			%			\FOR {$t=1$ to $T$}
			%			\STATE Fetch mini-batch $D_t$ from $D$
			%			\FOR {$i=1$ to $C-1$ } 
			%			\STATE Calculate $D^{s}_{i,t}$ using $\sigma_m$  by Eq.(\ref{data^s});
			%			\ENDFOR
			%			\STATE Calculate $D^s_{u,t}$ using  $\sigma_u $ by Eq.(\ref{data_u});
			%			\STATE Calculate  risk $\hat R_{SEA}(f_1,...,f_C)$ on sieved datasets $D^{s}_{1,t}$,...,$D^{s}_{C-1,t}$ and $D^s_{u,t}$;
			%			\STATE Update $w$ = $O(\hat R_{SEA}, w)$
			%			\ENDFOR
			%			\STATE \bfseries Return $w$
			%		\end{algorithmic}
		%	\end{algorithm}
	\section{Theoretical Analysis}
	In this section, we first introduce the  consistency analyses of the SEA risk $R_{SEA}$. Then, we show the generalization error bounds for the proposed method.
	\subsection{Consistency Analyses} 
	Fix $\bm f$, according to Eq.(\ref{data^s}) and Eq.(\ref{data_u}), we can partition all possible $D_m$ and $D_u$ into ${\mathscr{D}^s}(\bm f) = \{ (D^s_m, D^s_u)\} $ and  $\overline {\mathscr{D}^s}(\bm f) =  \{ (\overline D^s_m, \overline D^s_u)\} $, where $\overline D^s_m = \{ ({x_i},{y_i}) | CLm( \bm{f},(x_i,y_i))  < \sigma_m \}_{i=1}^{n_m - n_m^s}$ and $\overline D^s_u = \{ {x_i} |  CLu(\bm{f},x_i) < {\sigma _u}\}_{i=1}^{n_u- n_u^s} $. Then, we first  show the  bias of SEA risk $\hat R_{SEA}$  is positive. 
	
	\noindent
	\textbf{Lemma 2.} If the probability measure of $ {\mathscr{D}^s}(\bm f)$ and $ { \overline{\mathscr{D}}^s}(\bm f)$ are non-zero, then we have 
	\begin{equation}
		\mathbb E_{(p_m, p_u)}[\hat R_{SEA}({\bm f})] - {R_{mpu}({\bm f})} > 0
	\end{equation}where $R_{mpu}({\bm f})$ denotes the  MPU risk.
	
	Lemma 2 analyzes the bias of the proposed method under probability measure of sieved examples. Then, we establish the bound for  probability measure  $ {\mathscr{D}^s}(\bm f)$, which show the exponential decay of bias.
	
	\noindent
	\textbf{Lemma 3.} If the probability measure of $ {\mathscr{D}^s}(\bm f)$ and $ { \overline{\mathscr{D}}^s}(\bm f)$ are non-zero, and there  are $\alpha_m >0 $ and $\alpha_u >0 $, $\qquad$ such that $ \mathbb E_{p_m}[CLm(\bm{f},(x,y))] \leq \sigma_m - \alpha_m $ and $\mathbb E_{p_u}[CLu(\bm{f},x)] \leq \sigma_u - \alpha_u $. Let $C_m>0$, $C_u>0$ and $CLm(\bm{f},(x,y)) \leq C_m$, $CLu(\bm{f},x)\leq C_u$, the probability measure of  $ {\mathscr{D}^s}(\bm f)$ can be bounded by 
	\begin{equation}\label{prd}
		\begin{split}
			\Pr ( {\mathscr{D}^s}(\bm f)) \le \exp \big( - 2\big(\alpha _m^2 n_m^s C_m^2 + \alpha _u^2 n_u^s C_u^2\big)/(C_m^2C_u^2)\big)
		\end{split}
	\end{equation} where $n_m^s$ and $n_u^s$ denote the number of multi-positive and unlabeled examples.
	
	Lemma 3 shows the probability measure $ {\mathscr{D}^s}(\bm f)$ upper bounds. Then, by using Lemma 3, we can obtain the infinite-sample consistency for the proposed ESA risk.
	
	\noindent
	\textbf{Theorem 4.} Assume that there  are $\alpha_m >0 $ and $\alpha_u >0 $, such that $\mathbb E_{p_m}[CLm(\bm{f},(x,y))] \leq \sigma_m - \alpha_m$ and $\mathbb E_{p_u}[CLu(\bm{f},x)] \leq \sigma_u - \alpha_u$ and ${\Delta_{\bm f}}$ denotes the right-hand side of Eq.(\ref{prd}), $ \kappa_{C_l}^{\pi} = (C-1)\pi^{*}n_{m}^{*}C_{m} + C_{u}n_{u}^{s}$. As $n_m, n_u \rightarrow \infty$, the bias of $\hat R_{SEA}$ decays exponentially as follows:
	\begin{equation}
		\mathbb E_{(p_m, p_u)}[\hat R_{SEA}({\bm f})] - {R_{mpu}({\bm f})}< \kappa_{C_l}^{\pi} {\Delta_{\bm f}}
	\end{equation}
	
	Moreover, for any $\delta >0$, and for any $x\in \mathbb{R}^d$, $sup_{x}\phi(x)\leqslant C_{\phi}$, let  ${\chi _{{n_m},{n_u}}}={C_\phi }(\sum\nolimits_{i = 1}^{C - 1} {2{\pi _i}} \sqrt {{2 \mathord{\left/
				{\vphantom {2 {{n_i}\log {2 \mathord{\left/
									{\vphantom {2 \delta }} \right.
									\kern-\nulldelimiterspace} \delta }}}} \right.
				\kern-\nulldelimiterspace} {{n_i}\log {2 \mathord{\left/
						{\vphantom {2 \delta }} \right.
						\kern-\nulldelimiterspace} \delta }}}}  + \sqrt {{2 \mathord{\left/
				{\vphantom {2 {{n_u}\log {2 \mathord{\left/
									{\vphantom {2 \delta }} \right.
									\kern-\nulldelimiterspace} \delta }}}} \right.
				\kern-\nulldelimiterspace} {{n_u}\log {2 \mathord{\left/
						{\vphantom {2 \delta }} \right.
						\kern-\nulldelimiterspace} \delta }}}} )$, with probability at least $1-\delta/2$, 
	\begin{equation}
		\left| {\hat R_{SEA}({\bm f}) - R_{mpu}({\bm f})} \right| \le {\chi _{{n_m},{n_u}}} + \kappa_{C_l}^{\pi} {\Delta_{\bm f}}
	\end{equation} where $C^{s}$ denotes the upper bound,  $\pi^* = \mathop {\max }\limits_i \pi_i$, $n_{m}^{*} = \mathop {\min }\limits_i n_{i}^{s}$.
	
	Theorem 4 shows that with an increasing number of multi-positive and unlabeled data, the empirical risk of SEA method $\hat R_{SEA}({\bm f})$ converges to the multi-positive and unlabeled learning risk $R_{mpu}({\bm f})$. Then, from Theorem 4, we can find that the proposed ESA  is consistent for fixed $\bm f$.  Thus, the proposed ESA  is a biased yet consistent estimator to the risk. 
	
	\subsection{Generalization Error Bound}
	\begin{table*}
		\centering
		\caption{Classification accuracy of each algorithm on benchmark datasets, with varying classes. (MP), N means that classes $(1, 2,3), 4 $ are taken as multi-positive and negative classes respectively. R denotes a class randomly  selected from the datasets.   We report the mean and standard deviation of results over 5 trials. The best method is shown in bold (under 5$\%$ t-test). }
		\label{tab_acc_mnist}
		\renewcommand\arraystretch{1.2}
		\begin{tabular*}{\textwidth}{@{\extracolsep{\fill}}lc|ccc ccc }
			\toprule
			Dataset&(MP), N&UPU&NNPU&MPU&NMPU&CoMPU&ESA\\
			\cmidrule{1-8} 
			&(1, 2, 3), 0&50.78 $\pm$ 0.05&50.80 $\pm$ 1.10&97.89 $\pm$ 0.50&98.03 $\pm$ 0.55&98.01 $\pm$ 0.32&\textbf{99.30 $\pm$ 0.12}\\ 
			&(1, 2, 3), 5&49.57 $\pm$ 0.07&48.75 $\pm$ 0.27&96.91 $\pm$ 0.51&97.96 $\pm$ 0.28&97.37 $\pm$ 0.10&\textbf{99.33 $\pm$ 0.07}\\ 
			MNIST&(1, 2, 3), 9&50.87 $\pm$ 0.17&49.91 $\pm$ 0.39&97.26 $\pm$ 0.25&97.57 $\pm$ 0.34&97.34 $\pm$ 0.25&\textbf{99.06 $\pm$ 0.29}\\ 
			&(4, 5, 6), 2&51.56 $\pm$ 0.10&49.14 $\pm$ 2.41&97.49 $\pm$ 0.54&98.06 $\pm$ 0.25&97.98 $\pm$ 0.22&\textbf{99.12 $\pm$ 0.02}\\
			&(R, R, R), R&51.28 $\pm$ 0.24&46.21 $\pm$ 1.48&97.25 $\pm$ 0.29&97.59 $\pm$ 0.65&97.78 $\pm$ 0.18&\textbf{99.27 $\pm$ 0.04}\\
			%&(6,7,8,9), 0&&&&&\textbf{0}\\ 
			%	&(7,8,9), 1, 2 &&&&&\textbf{0}\\ 
			\cmidrule{1-8} 				
			&(1, 2, 3), 0&48.65 $\pm$ 0.31&47.33 $\pm$ 0.69&93.70 $\pm$ 0.20&93.71 $\pm$ 0.35&93.45 $\pm$ 0.21&\textbf{95.01 $\pm$ 0.08}\\ 
			&(1, 2, 3), 5&49.98 $\pm$ 0.01&49.57 $\pm$ 0.06&96.83 $\pm$ 0.19&96.91 $\pm$ 0.13&96.28 $\pm$ 0.17&\textbf{97.75 $\pm$ 0.08}\\
			%&5,3,1&8339&9955&43.48 $\pm$ 0.40&41.24 $\pm$ 5.40&91.22 $\pm$ 0.47&69.25 $\pm$ 2.66&\textbf{98.92 $\pm$ 0.21}\\ 
			Fashion&(1, 2, 3), 9&49.98 $\pm$ 0.01&46.52 $\pm$ 5.24&96.90 $\pm$ 0.32&96.98 $\pm$ 0.03&96.32 $\pm$ 0.22&\textbf{97.85 $\pm$ 0.19}\\ 
			%&2,4,6&9124&8594&32.34 $\pm$ 1.11&34.91 $\pm$ 5.97&91.90 $\pm$ 0.62&61.92 $\pm$ 0.68&\textbf{99.00 $\pm$ 0.21}\\
			&(4, 5, 6), 2&43.04 $\pm$ 0.42&34.41 $\pm$ 1.77&81.20 $\pm$ 1.62&79.79 $\pm$ 1.65&80.34 $\pm$ 0.24&\textbf{87.09 $\pm$ 0.61}\\
			&(R, R, R), R&49.88 $\pm$ 0.06&49.75 $\pm$ 0.28&90.34 $\pm$ 0.60&90.32 $\pm$ 1.11&90.12 $\pm$ 0.51&\textbf{92.95 $\pm$ 0.33}\\
			%	&(1,2,3,4), 0&&&&&\textbf{0}\\
			%&1,3,5&7033&10867&38.46 $\pm$ 6.44&41.60 $\pm$ 4.45&79.10 $\pm$ 1.43&60.66 $\pm$ 1.87&\textbf{98.16 $\pm$ 0.21}\\  
			%&(7,8,9), 1, 2&&&&&\textbf{0}\\
			\cmidrule{1-8} 				
			&(1, 2, 3), 0&48.50 $\pm$ 0.15&46.64 $\pm$ 0.86&87.79 $\pm$ 1.44&89.09 $\pm$ 2.20&88.83 $\pm$ 0.30&\textbf{94.20 $\pm$ 0.22}\\ 
			&(1, 2, 3), 5&45.40 $\pm$ 0.15&43.64 $\pm$ 0.13&84.90 $\pm$ 0.83&85.83 $\pm$ 1.38&85.02 $\pm$ 0.27&\textbf{90.85 $\pm$ 0.52}\\
			&(1, 2, 3), 9&46.63 $\pm$ 0.24&42.45 $\pm$ 1.12&85.01 $\pm$ 1.32&86.23 $\pm$ 1.32&85.35 $\pm$ 0.02&\textbf{93.09 $\pm$ 0.22}\\ 
			Kuzushi&(4, 5, 6), 2&45.87 $\pm$ 0.13&41.33 $\pm$ 0.40&82.10 $\pm$ 0.63&83.15 $\pm$ 1.63&82.47 $\pm$ 0.05&\textbf{90.50 $\pm$ 0.70}\\
			&(R, R, R), R&44.68 $\pm$ 0.14&38.73 $\pm$ 3.06&83.75 $\pm$ 1.04&84.42 $\pm$ 0.57&84.05 $\pm$ 0.27&\textbf{91.61 $\pm$ 0.77}\\
			&(R, R, R), R&46.24 $\pm$ 0.09&40.46 $\pm$ 2.52&86.40 $\pm$ 0.78&84.02 $\pm$ 3.26&85.37 $\pm$ 0.23&\textbf{91.77 $\pm$ 0.19}\\ 
			&(R, R, R), R&46.78 $\pm$ 0.47&43.92 $\pm$ 1.10&81.77 $\pm$ 2.50&83.29 $\pm$ 2.14&82.47 $\pm$ 0.70&\textbf{91.93 $\pm$ 0.80}\\
			\bottomrule
		\end{tabular*}
	\end{table*}
	
	In this section, we introduce the generalization error bound for the proposed ESA implemented by deep models with the OVR strategy. Let $\bm f = (f_1,...,f_C)$ be the classifier vector function in hypothesis set $\mathcal{H}$ of deep neural networks. Assume the surrogate loss $sup_{z}\phi(z)\leqslant C_{\phi}$, for $C_{\phi}>0$ and $L_{\phi}$ be the Lipschitz constant of $\phi$. By using Rademacher complexity\cite{DBLP:books/daglib/0034861,DBLP:books/daglib/0033642}, we will derive the following lemma.
	
	\noindent
	\textbf{Lemma 5.} For any $\delta>0$, with the probability at least $1-\delta/2$, we have
	\begin{equation}
		\begin{split}
			{\sup _{\bm f \in \mathcal{H}}}| {{R_{i}}(\bm f)-  {\widehat{R}_{i}}  (\bm f)} | \leqslant  8{N^s}{L_\phi }{\mathfrak{R}_{{n_{i}}}}(\mathcal{H}) + 2{\pi_{*}}{C_\phi }\sqrt {\frac{{2\ln (2/\delta )}}{{{n_{i}}}}} 
		\end{split}
	\end{equation}
	\begin{equation}
		\begin{split}
			{\sup_{\bm f \in \mathcal{H}}}|{{R_{u}}(\bm f)-  {\widehat{R}_{u}}  (\bm f)} | \leqslant  4{N^s}{L_\phi }{C}{\mathfrak{R}_{{n_{u}}}}(\mathcal{H}) + 2{C_\phi}{C}\sqrt {\frac{{2\ln (2/\delta )}}{{{n_{u}}}}} 
		\end{split}
	\end{equation}where $N^s = \mathop {\max }\limits_i ({n_i}/{n_i^s})$,  $C$ denotes number of classes, and 
	\begin{equation*}
		\begin{split}
			{R_{i}}(\bm f) = {\pi_i}\mathbb{E}_{p^s_i} \big[\phi\big({f_i}(x)\big)  - \phi\big({f_C}(x)\big) + \phi( - {f_C}(x)) - \phi( - {f_i}(x)) \big]
		\end{split}
	\end{equation*} denotes the proposed ESA risk of i-$th$ class and $\\ {{R_{u}}(\bm f) = \mathbb{E}_{p^s_u}}\big[\phi({f_C}(x)) + \sum\limits_{y \ne C} {\phi( - {f_y}(x))} \big]$ denotes the risk of unlabeled class, $\hat{R_{i}}(\bm f)$ and  ${\widehat{R}_u}  (\bm f)$ denote the empirical risk estimator to  $R_{i}(\bm f)$ and $R_u(\bm f)$ respectively, $\mathfrak{R}_{{n_{i}}}(\mathcal{H})$ and $\mathfrak{R}_{{n_u}}(\mathcal{H})$ are the Rademacher complexities of $\mathcal{H}$ for the sampling size $n_{i}$ from i-$th$ multi-positive data density and the sampling size $n_u$ from unlabeled data density.  
	
	According to Lemma 5, we can derive the following generalization error bound.
	
	\noindent
	\textbf{Theorem 6.} For any $\delta>0$, with the probability at least $1-\delta/2$, we have
	\begin{equation}
		\begin{split}
			R_{ESA}&({\hat {\bm f}_{ESA})} -  \mathop{\rm {min}} _{{\bm f} \in \mathcal{H}}R_{ESA}(\bm f) \leqslant  \\& \sum\nolimits_{i = 1}^{C - 1} {16{N^s}{L_\phi }{\mathfrak{R}_{{n_{i}}}}(\mathcal{H}) } + 8{N^s}{L_\phi }{C}{\mathfrak{R}_{{n_{u}}}}(\mathcal{H})\\&+  \sum\nolimits_{i = 1}^{C - 1}{4{\pi_{*}}{C_\phi }\sqrt {\frac{{2\ln (2/\delta )}}{{{n_{i}}}}} } + 4{C_\phi }{C}\sqrt {\frac{{2\ln (2/\delta )}}{{{n_{u}}}}} 
		\end{split}
	\end{equation} where $\hat {\bm f}_{ESA}$ denotes the trained model by minimizing the ESA risk $\hat{R}_{ESA}  (\bm f)$.
	
	Theorem 6 shows that the proposed ESA estimator exists an error bound and can achieve the optimal convergence rate. It is obvious that the error decreases when the number of multi-positive and unlabeled data grows. If the hypothesis set $\mathcal{H}$ of deep neural networks is fixed and the Rademacher complexity  $\mathfrak{R}_{{n}} \leqslant C_{{\bm f}}/\sqrt{n}$, we have $\mathfrak{R}_{{n_{i}}}(\mathcal{H}) = \mathcal{O}(1/\sqrt{n_{i}})$ and  $\mathfrak{R}_{{n_u}}(\mathcal{H}) = \mathcal{O}(1/\sqrt{n_u})$\cite{DBLP:journals/chinaf/GaoZ16}, then 
	\begin{equation*}
		\begin{split}
			{n_{1}},..., n_{C-1}, {n_u} \to \infty  \Longrightarrow R_{ESA}({\hat {\bm f}_{ESA})} - \mathop{\rm {min}} _{{\bm f} \in \mathcal{H}}R_{ESA}(\bm f)  \to 0
		\end{split}
	\end{equation*} where $C_{{\bm f}}$ denotes the constant for network weight and feature norms.

	%\begin{table}
	%	\centering
	%	\vspace{0em}
	%	\caption{Classification accuracy of negative class on benchmark datasets, with varying classes and the number of examples. (MP), N means that classes $(1, 2, 3), 4 $ of benchmark dataset are taken as multi-positive and negative classes respectively.  We report the mean and standard deviation of results over 5 trials. The best method is shown in bold (under 5$\%$ t-test). }
	%	{\resizebox{0.48\textwidth}{30mm}{
			%			\renewcommand\arraystretch{1.1}
			%			\tabcolsep=0.2cm
			%			\begin{tabular}{lc|ccc }
				%				\toprule
				%				Dataset&(MP), N&MPU&NMPU&ESA\\
				%%				\cmidrule{1-5} 
				%%				&(1,2,3), 0&&&\textbf{99.12$\pm$0.16}\\ 
				%%				&(1,2,3), 5&&&\textbf{0}\\
				%%				MNIST&(1,2,3), 9&&&\textbf{0}\\ 
				%%				&(4,5,6), 2&&&\textbf{0}\\
				%%				&(4,5,6), 3&&&\textbf{0}\\
				%%				%				&(4,5,6), 8&&&\textbf{}\\
				%%				%				%&1,3,5&7033&10867&38.46$\pm$6.44&41.60$\pm$4.45&79.10$\pm$1.43&60.66$\pm$1.87&\textbf{98.16$\pm$0.21}\\  
				%%				%				&(7,8,9), 1&&&\textbf{}\\ 
				%				\cmidrule{1-5} 				
				%				&(1,2,3), 0&97.80$\pm$0.45&&\textbf{0}\\ 
				%				&(1,2,3), 5&&&\textbf{0}\\
				%				&(1,2,3), 9&&&\textbf{0}\\ 
				%				Kuzushi&(4,5,6), 2&&&\textbf{0}\\
				%				&(4,5,6), 3&&&\textbf{0}\\
				%				&(7,8,9), 2&&&\textbf{0}\\
				%				&(7,8,9), 3&&&\textbf{0}\\
				%				\cmidrule{1-5} 				
				%				&(1,2,3), 0&&&\textbf{}\\ 
				%				CIFAR-10 &(1,2,3), 5&&&\textbf{}\\
				%				&(1,2,3), 9&&&\textbf{}\\
				%				\bottomrule
				%	\end{tabular}}}
	%%	\vspace{-2em}
	%	\label{tab_iac_mnist}	
	%\end{table}

	\section{Experiments}
	In this section, we report experimental results from four aspects of the proposed ESA: 1. performance of classifying on multi-positive and negative classes; 2. Robust  performance of  inaccurate  training class priors; 3. Sensitivity of lower bounds; 4. Issue of Class Probabilities Shift. In this paper, classifiers are trained with multi-positive and unlabeled data, and the parameters of neural networks for  ESA and comparing methods are the same.
	
	\begin{table*}[!htbp]
		\centering
		\caption{Classification accuracy of each algorithm on  CIFAR-10 and CIFAR-100 datasets, with varying classes. (MP), N means that classes $(1, 2,3), 4 $ are taken as multi-positive and negative classes respectively. R denotes a class randomly  selected from the datasets.  We report the mean and standard deviation of results over 5 trials. The best method is shown in bold (under 5$\%$ t-test). }
		\renewcommand\arraystretch{1.2}
		\begin{tabular*}{\textwidth}{@{\extracolsep{\fill}}lc|cccc}
			\toprule
			Dataset&(MP), N&MPU&NMPU&CoMPU&ESA\\
			\cmidrule{1-6} 
			&(1, 2, 3), 0&73.19 $\pm$ 0.33&74.60 $\pm$ 1.35&71.71 $\pm$ 0.62&\textbf{78.66 $\pm$ 0.57}\\ 
			&(1, 2, 3), 8&75.56 $\pm$ 0.36&77.46 $\pm$ 0.60&74.15 $\pm$ 0.23&\textbf{81.47 $\pm$ 0.46}\\ 
			&(4, 5, 6), 9&76.15 $\pm$ 1.42&78.24 $\pm$ 0.56&75.30 $\pm$ 1.81&\textbf{81.83 $\pm$ 0.25}\\
			CIFAR10&(R, R, R), R&75.85 $\pm$ 0.23&78.64 $\pm$ 0.35&75.45 $\pm$ 1.21&\textbf{81.27 $\pm$ 0.71}\\
			&(1, 2, 3, 4, 5, 6, 7, 8, 9), 0&50.12 $\pm$ 1.34&53.23 $\pm$ 0.33&56.43 $\pm$ 0.91&\textbf{64.13 $\pm$ 0.10}\\
			&(0, 1, 2, 3, 4, 6, 7, 8, 9), 5&51.35 $\pm$ 0.29&52.83 $\pm$ 0.65&55.48 $\pm$ 0.29&\textbf{65.26 $\pm$ 0.29}\\
			&(0, 1, 2, 3, 4, 5, 6, 7, 8), 9&50.27 $\pm$ 1.68&54.63 $\pm$ 0.81&57.78 $\pm$ 0.38&\textbf{64.47 $\pm$ 0.34}\\
			&(R, R, R, R, R, R, R, R, R), R&50.65 $\pm$ 0.19&53.59 $\pm$ 0.35&56.88 $\pm$ 0.52&\textbf{65.07 $\pm$ 0.24}\\
			\cmidrule{1-6}
			&(1, 2, ... , 99),100&52.79 $\pm$ 0.63&55.63 $\pm$ 0.65&55.18 $\pm$ 0.32&\textbf{61.24 $\pm$ 0.37}\\
			CIFAR100&(1, ... , 49, 51, ... , 100), 50&55.25 $\pm$ 0.29&52.83 $\pm$ 0.65&56.42 $\pm$ 0.03&\textbf{62.16 $\pm$ 0.09}\\
			&(R, ... , R), R&54.31 $\pm$ 0.82&53.43 $\pm$ 0.32&55.48 $\pm$ 0.29&\textbf{61.46 $\pm$ 0.21}\\
			\bottomrule
		\end{tabular*}
		\label{table_acc_cifar}	
	\end{table*}
	
	\begin{figure}
		\centering
		\includegraphics[width=0.5\textwidth]{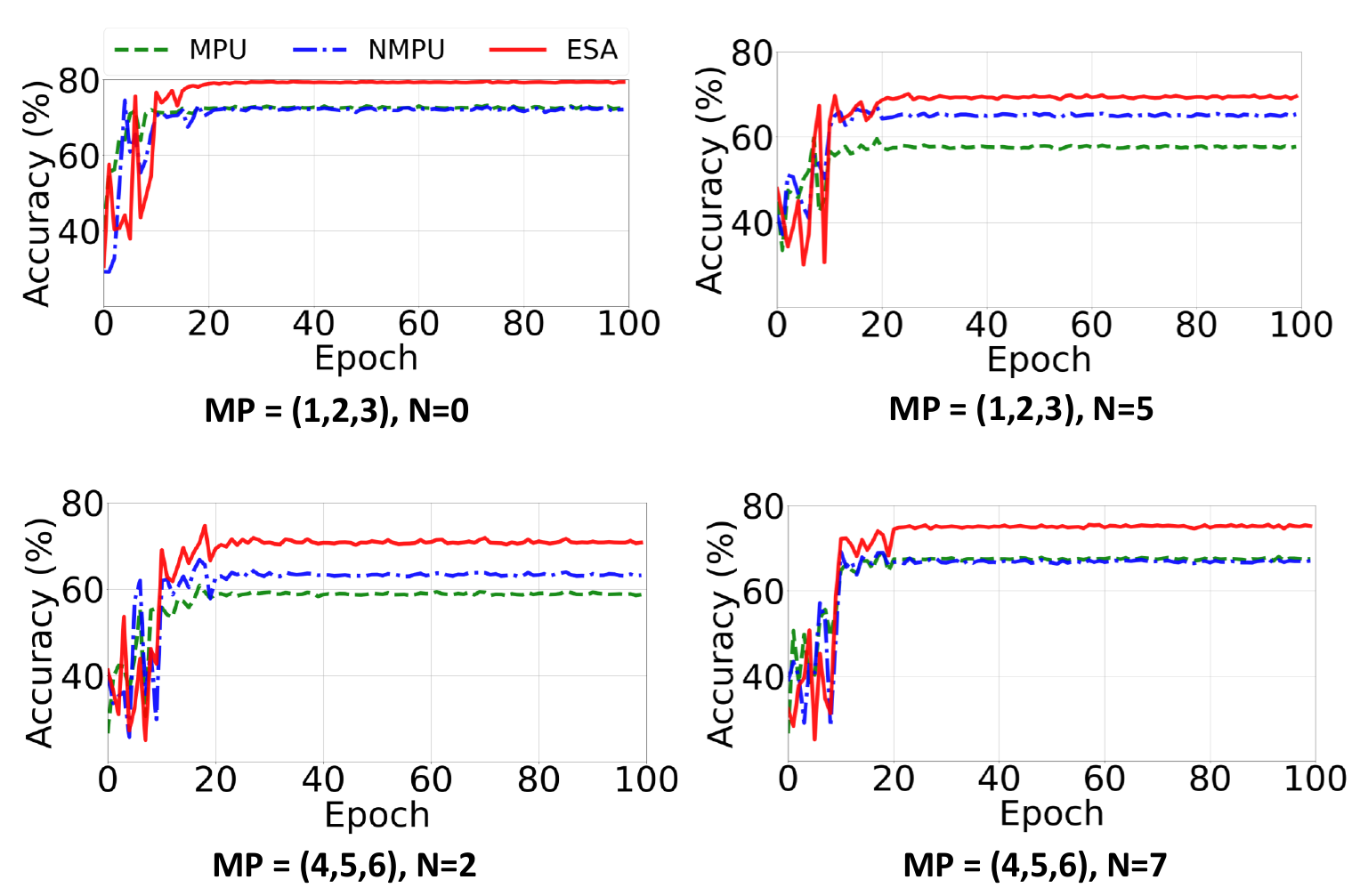}
		\caption{Illustrations test classification accuracy for all classes  on benchmark dataset CIFAR-10 in the training stage. MP $=(1, 2, 3)$ means that classes 1, 2, 3 are taken as multi-positive class. N $=0$ means that class 0 is taken as negative class. }
		\label{results_cifar}
	\end{figure}
	\subsection{Experiments Setup}
	\textbf{Datasets:} We train our deep model on four benchmark datasets: MNIST, Fashion-MNIST, Kuzushi-MNIST and CIFAR-10. Each dataset have 10 classes originally, and we constructed the multi-positive and unlabeled data from them as follows.  We first randomly selected some classes as multi-positive and another class as a negative class. Then, we randomly selected some examples from multi-positive classes as unlabeled and specified all of the negative examples as  unlabeled data.
	
	\noindent
	\textbf{Comparison Methods:} We absorb five state-of-the-art methods in two domains to evaluate  the  performance of proposed ESA. Two contends in the first group are methods of learning from multi-positive and unlabeled data (MPU\cite{DBLP:conf/ijcai/XuX0T17},  NMPU\cite{DBLP:conf/icdm/ShuLY020} and CoMPU\cite{DBLP:conf/acl/ZhouLL22}). The other two contends in the second group are methods of learning from positive and unlabeled data (UPU\cite{DBLP:conf/icml/PlessisNS15} and NNPU\cite{DBLP:conf/nips/KiryoNPS17}). MPU,  NMPU and CoMPU are powerful multi-positive and  unlabeled methods, which obtain better experimental results.
	
	\noindent
	\textbf{Common Setup:} In order to learn from multi-positive and unlabeled data, we use the OVR strategy to train deep neural networks implemented by margin square loss $\phi (z) = (1-z)^2$. We use Adadelta as an optimizer with initial an learning rate $8e^{-2}$ for MNIST, Fashion-MNIST, Kuzushi-MNIST and $5e^{-1}$ for CIFAR-10. For MNIST, Fashion-MNIST, Kuzushi-MNIST, we use a neural network with the same parameters for all methods which has two convolutional layers and two fully-connected layers.  For CIFAR-10 dataset, we also use the four convolutional layers and two fully-connected layers neural network for all methods to train classifiers.
	For CIFAR-100 dataset, we conduct experiments based on ResNet-18 with the same parameters for all methods. 
	For UPU and NNPU,  multi-positive dataset is treated as positive dataset. In this paper, all the experiments are conducted on PyTorch \cite{DBLP:conf/nips/PaszkeGMLBCKLGA19} and implementation on NVIDIA 3080Ti GPU.
	
	\subsection{Comparison with State-of-the-art Methods}
	The experimental results are reported in Table \ref{tab_acc_mnist}, where means and standard deviation of test accuracy of 5 trials are shown. The proposed ESA algorithm significantly outperforms by sieving out some examples, and achieves the best experimental results among all the approaches on four benchmark datasets. Then, the results of  NMPU (i.e., the non-negative risk estimator) are better than MPU, which accords with our discussion on the overfitting issue in the introduction. In addition, note that UPU and NNPU are trained on positive and unlabeled data. Hence UPU and NNPU only classify two of all classes in our experiments. The advantage of ESA motivates us to use more certain loss to sieve out some overfitting examples. Fig.\ref{results_cifar} illustrates the performance of the proposed ESA in the training stage, where the x-axis denotes the epoch and y-axis is the classification accuracy of testing data. As shown, the proposed ESA  outperforms others and  MPU obtains the similar performance compared with NMPU.

	\begin{figure*}
		\centering
		\includegraphics[width=0.98\textwidth]{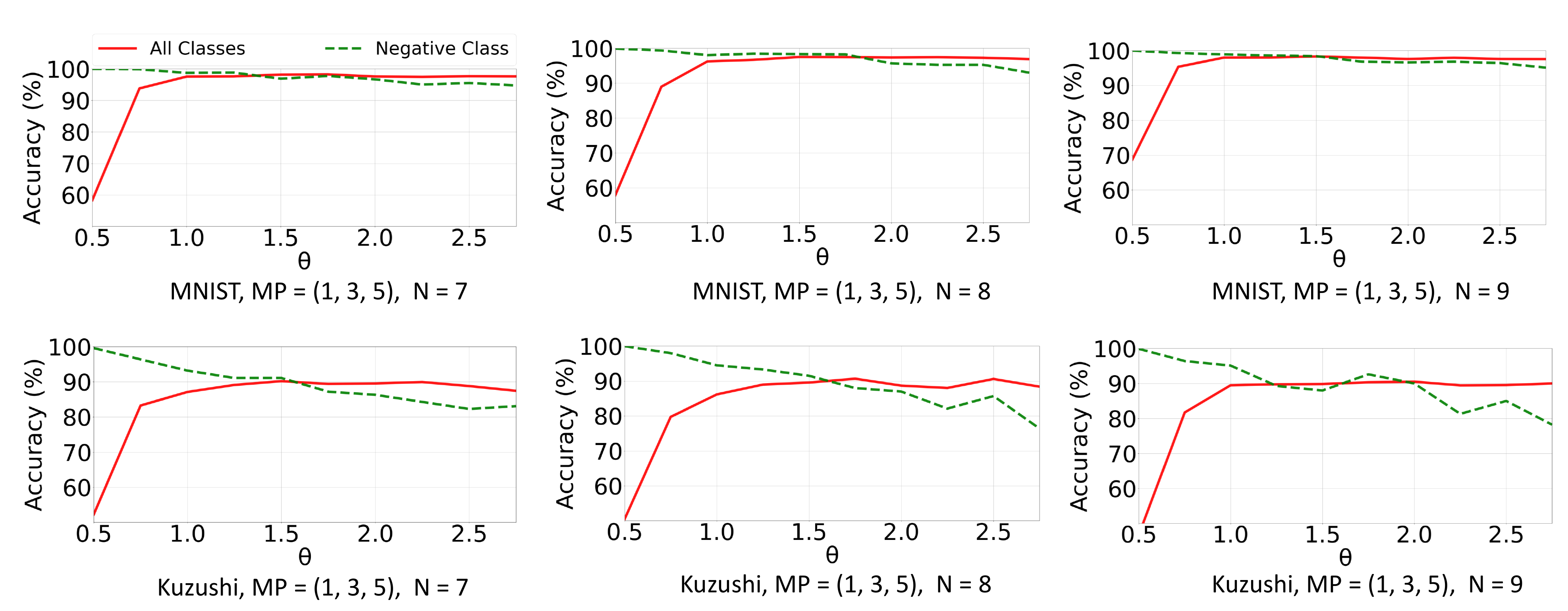}
		\caption{Illustrations classification accuracy for all classes and identifying accuracy for negative class  with various  perturbed mixture proportions. $\theta$ denotes the  perturbed rate for mixture proportion. MP $=(1, 2, 3)$ means that classes 1, 2, 3 are taken as multi-positive class. N $=4$ means that class 4 is taken as negative class. }
		\label{results_rate}
	\end{figure*}
	\begin{figure*}
		\centering
		\includegraphics[width=0.98\textwidth]{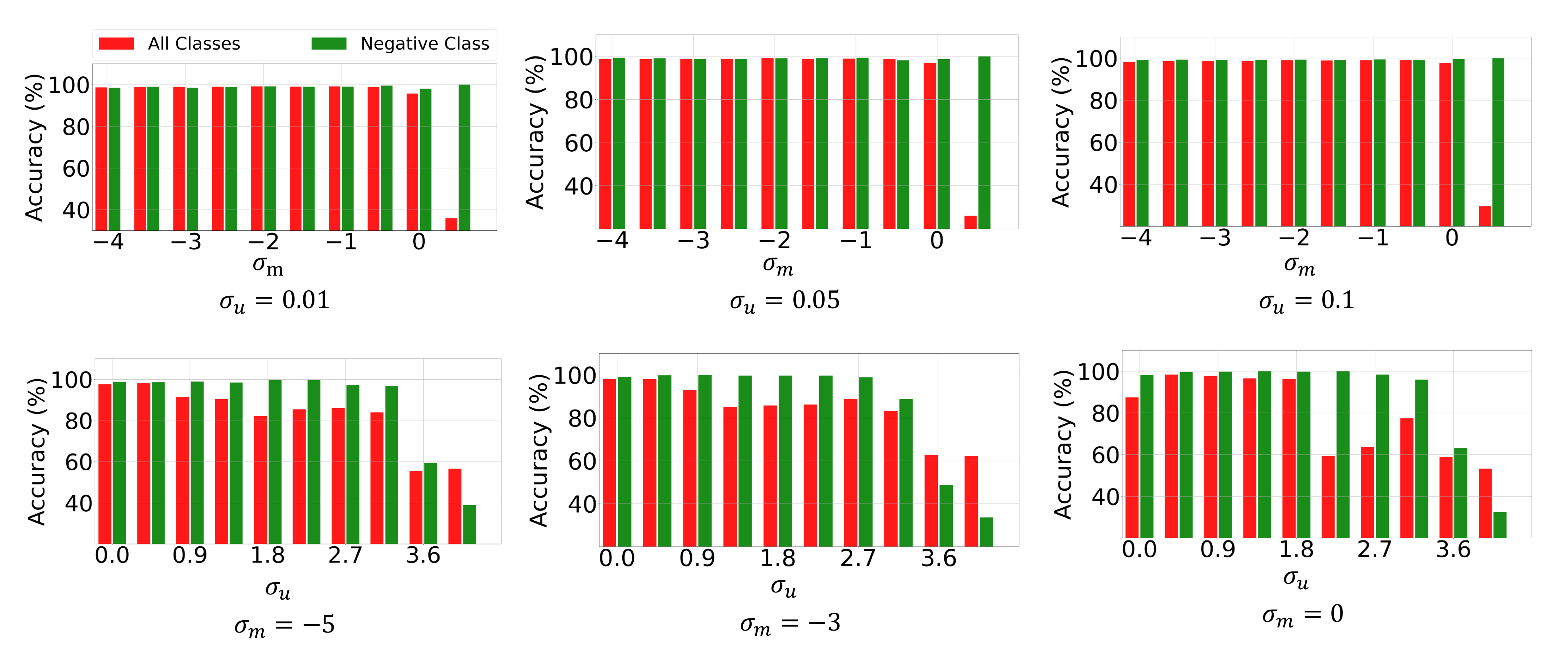}
		\caption{ Illustrations classification accuracy for all classes and identifying accuracy for negative class  with various  lower bounds. $\sigma_m$ and $\sigma_u $ denote the CL lower bounds for multi-positive and unlabeled data. In this experiment, the classes 1, 3, 5 are taken as multi-positive class,  and the class 7 is taken as negative class on benchmark dataset MNIST. }
		\label{results_below}
	\end{figure*}
	\begin{figure*}
		\centering
		\includegraphics[width=\textwidth]{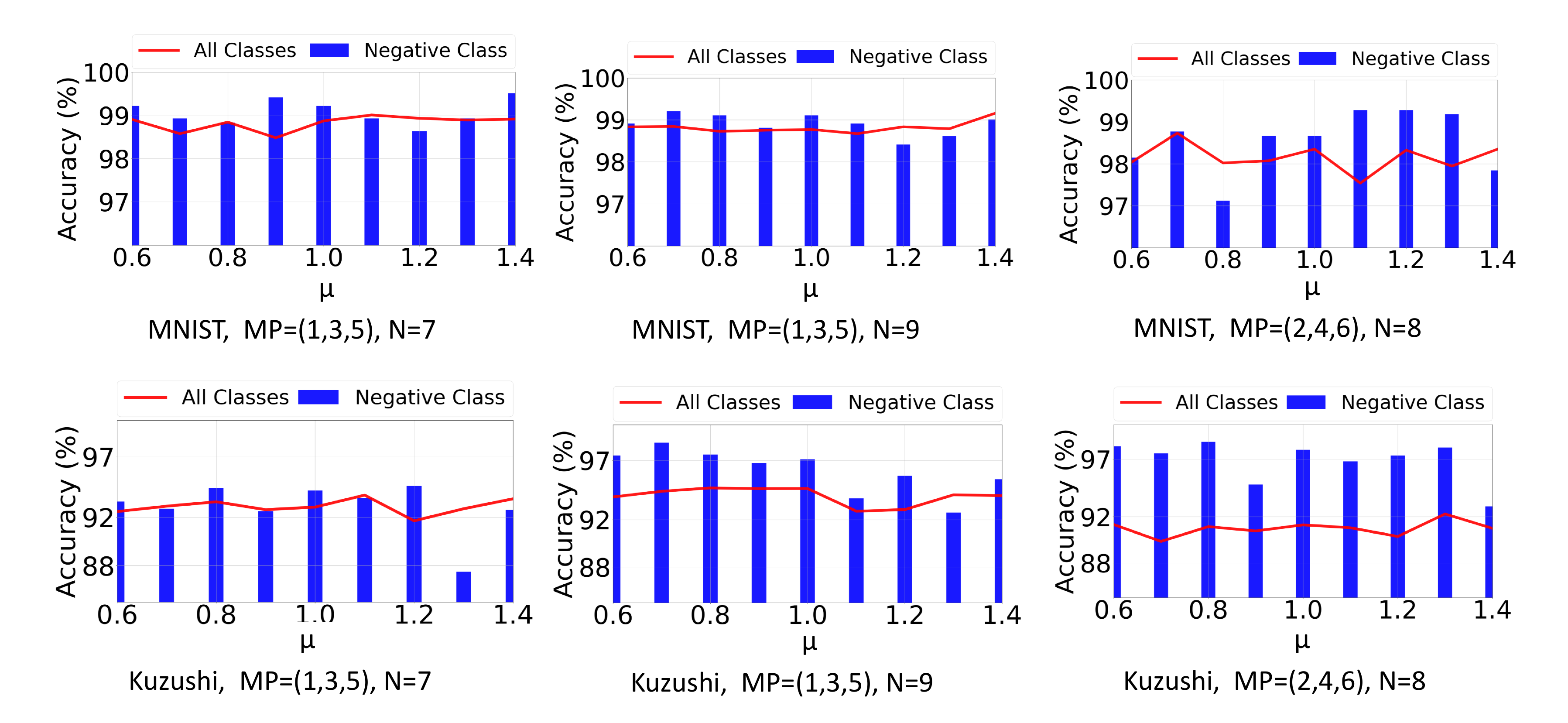}
		\caption{ Illustrations classification accuracy for all classes and identifying accuracy for negative class  with various test class distributions. $\mu$ denote the perturbed rate for test class distributions. MP $=(1, 3, 5)$ means that classes 1, 3, 5 are taken as multi-positive class. N $=7$ means that class 7 is taken as negative class. }
		\label{results_shf}
	\end{figure*}
	To further demonstrate the effectiveness and robustness of our method, we present more comprehensive experiments on CIFAR-10 and CIFAR-100 datasets. Specifically, for CIFAR-10 dataset, we add the case with more than 3 positive classes. For CIFAR-100 dataset, we add the case with 99 positive classes based on ResNet-18. The experimental results are reported in Table \ref{table_acc_cifar}, where means and standard deviation of test accuracy of 5 trials are shown. As can be seen from Table \ref{table_acc_cifar}, the proposed ESA algorithm achieves the best experimental results among all the approaches on four benchmark datasets.
	
	\subsection{Robustness for Inaccurate  Class Priors \label{ritc}}
	We evaluate the robustness of our algorithm for inaccurate train class priors on MNIST and Kuzushi-MNIST. Given the multi-positive and unlabeled data, the class priors (i.e., mixture proportion) can be estimated by some methods of Mixture Proportion estimation (MPE)\cite{DBLP:conf/icml/RamaswamyST16}. Without loss of generality, some experiments for evaluating the robustness of inaccurate train class priors are conducted by varying degrees of inaccuracies. Denote $\eta$ as a real number, $\pi' = \theta \pi$ as perturbed mixture proportions. In our experiments, the examples are sampled by $\pi$ but the classifiers are trained by using $\pi'$ instead. Fig.\ref{results_rate} shows the performance of the proposed ESA with various perturbed mixture proportions. We observe that when the mixture proportion varies from 0.75 to 2.5, the classifiers work well and avoid  overfitting in the testing datasets.  Then, the proposed ESA is robust to inaccurate mixture proportion, which means that we can use the surrogate class priors $\pi'(\pi<\pi'<2\pi)$ to train the classifiers. Besides, we found that using the surrogate class priors $\pi'$ with the bigger values will improve the generalization performance, which alleviates the overfitting problem.
	
	\subsection{Sensitivity of Lower Bounds}
	We evaluate the sensitivity of the lower bounds, which sieve some examples in the training stage. In our experiments, we fix $\sigma_m$ and select different $\sigma_u$ to train neural networks, and fix $\sigma_u$ to select $\sigma_m$. Fig.\ref{results_below} plots the performance of ESA  with different lower bounds, where y-axis is the testing accuracy and x-axis denotes the value of lower bounds. After inspecting the experimental result, we found that the performance of the proposed ESA will degrade when the lower bound is too large for training classifiers. These results indicate that an appropriate lower bound will improve the generalization performance.
	\subsection{Issue of Class Probabilities Shift}
	In the above setting, the distribution of unlabeled data is the same as the testing distribution. In this section, we evaluate the proposed ESA for investigating class distribution shift in the testing data.   Without loss of generality, we conduct the experiments on varying degrees of distribution shift for testing data. Let $\mu$ be a real number, $\pi'_{te} = \mu \pi$ be the perturbed distribution for testing data. In this section, the classifiers are trained by using $\pi$ while tested by $\pi'_{te}$. The experimental results on benchmark datasets MNIST and Kuzushi MNIST are shown in Fig.\ref{results_shf}. As we can see, the performance of proposed ESA is well on varying degrees of distribution shift, which is clearly in accordance with theoretical analysis, i.e., the proposed ESA is consistent.
	
	\section{Conclusion}
	In this paper, we investigate the problem  of  minimum risk shifting for multi-positive and unlabeled learning in the training stage. We propose an Example Sieve Approach (ESA) to select examples for training multi-class classifier to alleviate the severe problem  of minimum risk shifting. Specifically, we sieve out some examples by utilizing the loss values of each example  in the training stage, and analyze the consistency of proposed risk estimator. Besides, we show that the  estimation error of proposed ESA obtains the optimal parametric convergence rate. Extensive experiments on various real-world datasets show the proposed approach outperforms previous methods.
	
	In the future, we will investigate an advanced sieve approach, which not only sieve out some overfitting examples, but also is independent of loss value of each example. Besides, it would be interesting to study the sieve approach for other weakly-supervised learning.
	
	\section{Acknowledgments}
	This work was supported by the National Natural Science Foundation of China (No. 61976217, 62306320), the Open Project Program of State Key Lab. for Novel Software Technology (No. KFKT2024B32), the Natural Science Foundation of Jiangsu Province (No. BK20231063).
	
%	the Graduate Innovation Program of China University of Mining and Technology (No. 2024WLKXJ188, 2024WLJCRCZL262), and the Postgraduate Research $\&$ Practice Innovation Program of Jiangsu Province (No. KYCX24$\_$2785)

\bibliographystyle{ACM-Reference-Format}
\bibliography{sample-base}

\appendix
\section{Proof of Proposition 1}
\textbf{Proposition 1.} Let  $ p^s_i = p^s(x|y=i)$ denotes aconditional density of sieved dataset, then the risk of example sieve can be equivalently represented as:
\begin{equation}\label{sre}
	\begin{split}
		R_{ESA}&(f_1,...,f_C) = \sum\limits_{i = 1}^{C - 1} {{\pi _i}{\mathbb{E}_{p^s_i}}\bigg[\phi\big({f_i}(x)\big)}  - \phi\big({f_C}(x)\big) \\& + \big[\phi( - {f_C}(x)) - \phi( - {f_i}(x)\big]\bigg] \\& + {\mathbb{E}_{p^s_u}}\big[\phi({f_C}(x)) + \sum\limits_{y \ne C} {\phi( - {f_y}(x))} \big]
	\end{split}
\end{equation}
\emph{Proof.} According to the collecting procedure of multi-positive and unlabeled data, we have 
\begin{equation}\label{mpu}
	\begin{split}
		R_{mpu}(\bm{f}) = &\sum\limits_{i = 1}^{C - 1} {\pi _i}{\mathbb{E}_i}[L(\bm{f}(x),y = i)   - L(\bm{f}(x),y = C)] \\& + {\mathbb{E}_{p_u}}[L(\bm{f}(x),y = C)]
	\end{split}
\end{equation}

We adopt the one-versus-rest (OVR) strategy, then the risk can be represented as:
\begin{equation}\label{ovrmpu}
	\begin{split}
		R_{mpu}(\bm{f}) = &\sum\limits_{i = 1}^{C - 1} {\pi _i}{\mathbb{E}_i}[\phi(f_i(x))+\sum\limits_{y'=1,y' \ne i}^{y'=C} {\phi( - {f_{y'}}(x))}  \\& - (\phi(f_C(x))+\sum\limits_{y'=1}^{y'=C-1} {\phi( - {f_{y'}}(x))})] \\& + {\mathbb{E}_{p_u}}[\phi(f_C(x))+\sum\limits_{y'=1}^{y'=C-1} {\phi( - {f_{y'}}(x))}]
	\end{split}
\end{equation}

Hence, we have 
\begin{equation}\label{sre}
	\begin{split}
		R_{ESA}&(f_1,...,f_C) = \sum\limits_{i = 1}^{C - 1} {{\pi _i}{\mathbb{E}_{p_i}}\bigg[\phi\big({f_i}(x)\big)}  - \phi\big({f_C}(x)\big) \\& + \big[\phi( - {f_C}(x)) - \phi( - {f_i}(x)\big]\bigg] \\& + {\mathbb{E}_{p_u}}\big[\phi({f_C}(x)) + \sum\limits_{y \ne C} {\phi( - {f_y}(x))} \big]
	\end{split}
\end{equation}

Then, we use sieved densities instead of MPU densities, and the risk can be represented as:
\begin{equation}\label{sre}
	\begin{split}
		R_{ESA}&(f_1,...,f_C) = \sum\limits_{i = 1}^{C - 1} {{\pi _i}{\mathbb{E}_{p_i^s}}\bigg[\phi\big({f_i}(x)\big)}  - \phi\big({f_C}(x)\big) \\& + \big[\phi( - {f_C}(x)) - \phi( - {f_i}(x)\big]\bigg] \\& + {\mathbb{E}_{p_u^s}}\big[\phi({f_C}(x)) + \sum\limits_{y \ne C} {\phi( - {f_y}(x))} \big]
	\end{split}
\end{equation} which concludes the proof. \hfill $\square$

\section{Proof of Lemma 2}
\textbf{Lemma 2.} If the probability measure of $ {\mathscr{D}^s}(\bm f)$ and $ { \overline{\mathscr{D}}^s}(\bm f)$ are non-zero, then we have 
\begin{equation}
	\mathbb E_{(p_m, p_u)}[\hat R_{ESA}({\bm f})] - {R_{mpu}({\bm f})} > 0
\end{equation}where $R_{mpu}({\bm f})$ denotes the unbiased MPU risk.

\emph{Proof.} Let $D_m = \bigcup\limits_{i = 1}^{C - 1} {{D_i}} $ denotes the dataset of multi-positive classes,$p_m$ and $p_u$ denote the distribution function of dataset $D_m$ and $D^u$, $CLi(\bm f(x_k)) = [\phi({f_i}({x_k}))  - \phi( {f_C}({x_k}))  + \phi( - {f_C}({x_k})) - \phi({-f_i}({x_k})]$, we have 	\begin{equation}\begin{split}
		&\mathbb E_{(p_m, p_u)}[\hat R_{ESA}({\bm f})] - {R_{mpu}({\bm f})} \\ = &\int \hat R_{ESA}({\bm f}) -\hat R_{mpu}({\bm f}) d p_m p_u \\ = &\int \sum\limits_{i = 1}^{C - 1} \hat R_{i}({\bm f}) +  \hat R_{u}({\bm f}) -\sum\limits_{i = 1}^{C - 1} \hat R_{i,mpu}({\bm f}) +  \hat R_{u,mpu}({\bm f}) d p_m p_u\\ = &\int \sum\limits_{i = 1}^{C - 1} \underbrace {\hat R_{i}({\bm f}) - \hat R_{i,mpu}({\bm f})}_{\vartriangle } +  \hat R_{u}({\bm f}) - \hat R_{u,mpu}({\bm f})d p_m p_u
	\end{split}
\end{equation}
Then, if  $ { \overline{\mathscr{D}}^s}(\bm f)$ are non-zero,  $\vartriangle$ can be rewritten as 
\begin{equation*}
	\begin{split}
		\vartriangle =&  \frac{{{\pi _i}}}{{n_i^s}}\sum\limits_{{x_k} \in {D^s_i}} CLi(\bm f(x_k))  - \frac{{{\pi _i}}}{{n_i}}\sum\limits_{{x_k} \in {D_i}} CLi(\bm f(x_k)) \\=& \pi _i [\frac{{{1 }}}{{n_i^s}}\sum\limits_{{x_k} \in {D^s_i}} CLi(\bm f(x_k)) - \frac{{{1}}}{{n_i}}\sum\limits_{{x_k} \in {D^s_i}} CLi(\bm f(x_k)) \\&- \frac{{{1}}}{{n_i}}\sum\limits_{{x_k} \in {\overline D^s_i}} CLi(\bm f(x_k))]\\&>\pi _i [\frac{{{1 }}}{{n_i^s}}\sum\limits_{{x_k} \in {D^s_i}} CLi(\bm f(x_k)) - \frac{{{1}}}{{n_i}}\sum\limits_{{x_k} \in {D^s_i}} CLi(\bm f(x_k)) \\&- \frac{{{1}}}{{n_i}}(n_i- n_i^s) \sigma_m]\\=&  {(n_i-n_i^s) \mathord{\left/
				{\vphantom {n n}} \right.
				\kern-\nulldelimiterspace} n_i}*[\sigma_m' - \sigma_m] \geqslant  0
	\end{split}
\end{equation*}
where $\sigma_m' = min_{x_k\in D_i^s} CLi(\bm f(x_k)) $. 

$\hat R_{u}({\bm f}) - \hat R_{u,mpu}({\bm f})$ can be proven using the same proof technique, which proves Lemma 2.\hfill $\square$

\section{Proof of Lemma 3}

\textbf{Lemma 3.} If the probability measure of $ {\mathscr{D}^s}(\bm f)$ and $ { \overline{\mathscr{D}}^s}(\bm f)$ are non-zero, and there  are $\alpha_m >0 $ and $\alpha_u >0 $, such that $\mathbb E_{p_m}[CLm(\bm{f},(x,y))] \leq \sigma_m - \alpha_m$ and $\mathbb E_{p_u}[CLu(\bm{f},x)] \leq \sigma_u - \alpha_u$.Let $C_m>0$, $C_u>0$ and $CLm(\bm{f},(x,y)) \leq C_m$, $CLu(\bm{f},x)\leq C_u$, the probability measure of  $ {\mathscr{D}^s}(\bm f)$ can be bounded by 
\begin{equation}\label{prd}
	\begin{split}
		\Pr ( {\mathscr{D}^s}(\bm f)) \le &\exp \big( - 2\big(\alpha _m^2 n_m^s C_m^2 \\&+ \alpha _u^2 n_u^s C_u^2\big)/(C_m^2C_u^2)\big)
	\end{split}
\end{equation} where $n_m^s$ and $n_u^s$ denote the number of multi-positive and unlabeled examples.

\emph{Proof.} Let $Pr$ denotes the probability, we have 
\begin{equation*}
	\begin{split}
		Pr(\overline {\mathscr{D}^s}(\bm f)) = &Pr(\exists {x_i} \in {D_m},CLm(\bm f,(x,y)) < {\sigma _m} \\& or\; \exists {x_i} \in {D_u},CLu(\bm f,x) < {\sigma _u})\\& = 1- Pr(\forall {x_i} \in {D_m},CLm(\bm f,(x,y)) \geqslant {\sigma _m} \\& and\; \forall  {x_i} \in {D_u},CLu(\bm f,x) \geqslant {\sigma _u}) \\=& 1- \prod\limits_{i = 1}^{{n_m}} \underbrace{{\Pr (CLm(\bm f,(x,y)) \geqslant {\sigma _m})}}_{\diamondsuit} \\& * \prod\limits_{i = 1}^{{n_u}} {\Pr (CLu(\bm f,x) \geqslant {\sigma _u})} 
	\end{split}
\end{equation*}

Then, \begin{equation*}
	\begin{split}
		Pr&( {\mathscr{D}^s}(\bm f))  = 1- Pr(\overline {\mathscr{D}^s}(\bm f)) \\=&\prod\limits_{i = 1}^{{n_m}} \underbrace{{\Pr (CLm(\bm f,(x,y)) \geqslant {\sigma _m})}}_{\diamondsuit}   \prod\limits_{i = 1}^{{n_u}} {\Pr (CLu(\bm f,x) \geqslant {\sigma _u})} 
	\end{split}
\end{equation*}

If the  $\mathbb E_{p_m}[CLm(\bm{f},(x,y))] \leq \sigma_m - \alpha_m$, for  $\alpha_m >0 $, $\diamondsuit$ can be rewritten as follows:

\begin{equation*}
	\begin{split}
		\Pr (& CLm(\bm f,(x,y)) \geqslant {\sigma _m}) \\& \leq \Pr (CLm(\bm f,(x,y)) \geqslant {\mathbb E_{p_m}[CLm(\bm{f},(x,y))] + \alpha_m})\\&=\Pr (CLm(\bm f,(x,y)) - {\mathbb E_{p_m}[CLm(\bm{f},(x,y))] \geqslant \alpha_m})
	\end{split}
\end{equation*}

Then, according McDiarmid's inequality, we can obtain 
\begin{equation*}
	\begin{split}
		\Pr & (CLm(\bm f,(x,y)) - {\mathbb E_{p_m}[CLm(\bm{f},(x,y))] \geqslant \alpha_m}) \\& \leq exp({\raise0.7ex\hbox{${{\text{ - 2}}\alpha _m^2}$} \!\mathord{\left/
				{\vphantom {{{\text{ - 2}}\alpha _m^2} {C_m^2}}}\right.\kern-\nulldelimiterspace}
			\!\lower0.7ex\hbox{${C_m^2}$}}) 
	\end{split}
\end{equation*}

$\Pr (CLu(\bm f,x) \geqslant {\sigma _u})$ can be proven using the same proof technique, which proves Lemma 3.\hfill $\square$

\section{Proof of Theorem 4}

\textbf{Theorem 4.} Assume that there  are $\alpha_m >0 $ and $\alpha_u >0 $, such that $\mathbb E_{p_m}[CLm(\bm{f},(x,y))] \leq \sigma_m - \alpha_m$ and $\mathbb E_{p_u}[CLu(\bm{f},x)] \leq \sigma_u - \alpha_u$ and ${\Delta_{\bm f}}$ denotes the right-hand side of Eq.\ref{prd}, $ \kappa_{C_l}^{\pi} = (C-1)\pi^{*}n_{m}^{*}C_{m} + C_{u}n_{u}^{s}$. As $n_m, n_u \rightarrow \infty$, the bias of $\hat R_{ESA}$ decays exponentially as follows:
\begin{equation}
	\mathbb E_{(p_m, p_u)}[\hat R_{ESA}({\bm f})] - {R_{mpu}({\bm f})}< \kappa_{C_l}^{\pi} {\Delta_{\bm f}}
\end{equation}

Moreover, for any $\delta >0$, and for any $x\in \mathbb{R}^d$, $sup_{x}\phi(x)\leqslant C_{\phi}$, let  ${\chi _{{n_m},{n_u}}}={C_\phi }(\sum\nolimits_{i = 1}^{C - 1} {2{\pi _i}} \sqrt {{2 \mathord{\left/
			{\vphantom {2 {{n_i}\log {2 \mathord{\left/
								{\vphantom {2 \delta }} \right.
								\kern-\nulldelimiterspace} \delta }}}} \right.
			\kern-\nulldelimiterspace} {{n_i}\log {2 \mathord{\left/
					{\vphantom {2 \delta }} \right.
					\kern-\nulldelimiterspace} \delta }}}}  + \sqrt {{2 \mathord{\left/
			{\vphantom {2 {{n_u}\log {2 \mathord{\left/
								{\vphantom {2 \delta }} \right.
								\kern-\nulldelimiterspace} \delta }}}} \right.
			\kern-\nulldelimiterspace} {{n_u}\log {2 \mathord{\left/
					{\vphantom {2 \delta }} \right.
					\kern-\nulldelimiterspace} \delta }}}} )$, with probability at least $1-\delta/2$, 
\begin{equation}
	\left| {\hat R_{ESA}({\bm f}) - R_{mpu}({\bm f})} \right| \le {\chi _{{n_m},{n_u}}} + \kappa_{C_l}^{\pi} {\Delta_{\bm f}}
\end{equation} where $C^{s}$ denotes the upper bound,  $\pi^* = \mathop {\max }\limits_i \pi_i$, $n_{m}^{*} = \mathop {\min }\limits_i n_{i}^{s}$.

\emph{Proof.}  According to Lemma 2 and 3, the exponential decay of the biased is obtained via 
\begin{equation*}
	\begin{split}
		&\mathbb E_{(p_m, p_u)}[\hat R_{ESA}({\bm f})] - {R_{mpu}({\bm f})} \\=& \int \sum\limits_{i = 1}^{C - 1}[ \hat R_{i}({\bm f}) - \hat R_{i,mpu}({\bm f})] +  \hat R_{u}({\bm f}) - \hat R_{u,mpu}({\bm f})d p_m p_u \\=& \int_{ {\mathscr{D}^s}(\bm f))} \sum\limits_{i = 1}^{C - 1} [\hat R_{i}({\bm f}) - \hat R_{i,mpu,s}({\bm f})] +  \hat R_{u}({\bm f}) - \hat R_{u,mpu,s}({\bm f})d p_m p_u \\&- \int_{\overline {\mathscr{D}^s}(\bm f))} \sum\limits_{i = 1}^{C - 1}  \hat R_{i,mpu,\overline s}({\bm f}) +  \hat R_{u,mpu,\overline s}({\bm f})d p_m p_u \\ \leq & \int \sum\limits_{i = 1}^{C - 1} [\hat R_{i}({\bm f}) - \hat R_{i,mpu,s}({\bm f})] +  \hat R_{u}({\bm f}) - \hat R_{u,mpu,s}({\bm f})d p_m p_u  \\ \leq & sup \bigg[\sum\limits_{i = 1}^{C - 1} [\hat R_{i}({\bm f}) - \hat R_{i,mpu,s}({\bm f})] \\& +  \hat R_{u}({\bm f}) - \hat R_{u,mpu,s}({\bm f})\bigg] * \int_{{\mathscr{D}^s}(\bm f))} d p_m p_u\\ \leq & \kappa_{C_l}^{\pi} {\Delta_{\bm f}}
	\end{split}
\end{equation*}where $\hat R_{i,mpu,s}({\bm f}) = \frac{{{\pi _i}}}{{n_i}}\sum\limits_{{x_k} \in {D_i^s}} CLi(\bm f(x_k))$, $\hat R_{i,mpu,\overline s}({\bm f}) = \frac{{{\pi _i}}}{{n_i}}\sum\limits_{{x_k} \in {\overline D_i^s}} CLi(\bm f(x_k))$ and $\hat R_{u,mpu,s}({\bm f}) = \frac{{{1}}}{{n_u}}\sum\limits_{{x_k} \in {D_u^s}} CLu(\bm f(x_k))$, $\hat R_{u,mpu,\overline s}({\bm f}) = \frac{{{1}}}{{n_u}}\sum\limits_{{x_k} \in {\overline D_u^s}} CLu(\bm f(x_k))$

Then, 	the deviation bound is due to 

\begin{equation*}
	\begin{split}
		|\hat R_{ESA}({\bm f})& - R_{mpu}({\bm f}) | \leq | \hat R_{ESA}({\bm f}) - \mathbb E_{(p_m, p_u)} \hat R_{ESA}({\bm f})| \\&+ | \mathbb E_{(p_m, p_u)}[\hat R_{ESA}({\bm f})] - {R_{mpu}({\bm f})}| \\& \leq | \hat R_{ESA}({\bm f}) - \mathbb E_{(p_m, p_u)} \hat R_{ESA}({\bm f})| + \kappa_{C_l}^{\pi} {\Delta_{\bm f}} 
	\end{split}
\end{equation*}

According McDiarmid's inequality, we can obtain 
\begin{equation*}
	\begin{split}
		|& \hat R_{ESA}({\bm f}) - \mathbb E_{(p_m, p_u)} \hat R_{ESA}({\bm f})| \\& \leq {C_\phi }(\sum\nolimits_{i = 1}^{C - 1} {2{\pi _i}} \sqrt {{2 \mathord{\left/
					{\vphantom {2 {{n_i}\log {2 \mathord{\left/
										{\vphantom {2 \delta }} \right.
										\kern-\nulldelimiterspace} \delta }}}} \right.
					\kern-\nulldelimiterspace} {{n_i}\log {2 \mathord{\left/
							{\vphantom {2 \delta }} \right.
							\kern-\nulldelimiterspace} \delta }}}}   + \sqrt {{2 \mathord{\left/
					{\vphantom {2 {{n_u}\log {2 \mathord{\left/
										{\vphantom {2 \delta }} \right.
										\kern-\nulldelimiterspace} \delta }}}} \right.
					\kern-\nulldelimiterspace} {{n_u}\log {2 \mathord{\left/
							{\vphantom {2 \delta }} \right.
							\kern-\nulldelimiterspace} \delta }}}} )
	\end{split}
\end{equation*}which proves the whole theorem.\hfill $\square$

\section{Proof of Lemma 5}

\textbf{Lemma 5.} For any $\delta>0$, with the probability at least $1-\delta/2$, we have
\begin{equation}
	\begin{split}
		{\sup _{\bm f \in \mathcal{H}}}| & {{R_{i}}(\bm f)-  {\widehat{R}_{i}}  (\bm f)} | \\& \leqslant  8{N^s}{L_\phi }{\mathfrak{R}_{{n_{i}}}}(\mathcal{H}) + 2{\pi_{*}}{C_\phi }\sqrt {\frac{{2\ln (2/\delta )}}{{{n_{i}}}}} 
	\end{split}
\end{equation}
\begin{equation}
	\begin{split}
		{\sup _{\bm f \in \mathcal{H}}}| & {{R_{u}}(\bm f)-  {\widehat{R}_{u}}  (\bm f)} | \\& \leqslant  4{N^s}{L_\phi }{C}{\mathfrak{R}_{{n_{u}}}}(\mathcal{H}) + 2{C_\phi }{C}\sqrt {\frac{{2\ln (2/\delta )}}{{{n_{u}}}}} 
	\end{split}
\end{equation}where $N^s = \mathop {\max }\limits_i ({n_i}/{n_i^s})$,  $C$ denotes number of classes, and 
\begin{equation*}
	\begin{split}
		{R_{i}}(\bm f) = {\pi_i}\mathbb{E}_{p^s_i}& \big[\phi\big({f_i}(x)\big)  - \phi\big({f_C}(x)\big) \\& + \phi( - {f_C}(x)) - \phi( - {f_i}(x)) \big]
	\end{split}
\end{equation*} denotes the proposed ESA risk of i-$th$ class and ${{R_{u}}(\bm f) = \mathbb{E}_{p^s_u}}\big[\phi({f_C}(x)) + \sum\limits_{y \ne C} {\phi( - {f_y}(x))} \big]$ denotes the risk of unlabeled class, $\hat{R_{i}}(\bm f)$ and  ${\widehat{R}_u}  (\bm f)$ denote the empirical risk estimator to  $R_{i}(\bm f)$ and $R_u(\bm f)$ respectively, $\mathfrak{R}_{{n_{i}}}(\mathcal{H})$ and $\mathfrak{R}_{{n_u}}(\mathcal{H})$ are the Rademacher complexities of $\mathcal{H}$ for the sampling size $n_{i}$ from i-$th$ multi-positive data density and the sampling size $n_u$ from unlabeled data density. 

\emph{Proof.} Suppose the surrogate loss $\phi(z)$ is bounded by $su{p_{z}}\phi(z)\leqslant C_{\phi}$, let function $\Phi$ defined for any complementary labels  sample $S_{i}$ by $\Phi (S_{i}) = sup_{F\in \mathcal{H}} R_{i}( {\bm f}) -  \widehat R_{i}( {\bm f} )$. If  $x_i$ in complementary labels dataset is replaced with $x_i'$, the change of $\Phi_{i} (S_{i})$ does not exceed the supermum of the difference, we have
\begin{equation}
	\Phi_{i} (S_{i}') - \Phi_{i} (S_{i}) \leqslant \frac{{4\pi_ *{C_\phi }}}{{{n^s_{i}}}}
\end{equation} Then, by McDiarmid's inequality, the following holds: 
\begin{equation}
	\begin{split}
		P({\Phi _i}({S_i}) - E[{\Phi _i}({S_i})]  \geqslant& \varepsilon ) \leqslant \exp ( - {{2{\varepsilon ^2}} \mathord{\left/
				{\vphantom {{2{\varepsilon ^2}} {[{n_i}}}} \right.
				\kern-\nulldelimiterspace} {[{n_i}}}{\left( {{{4\pi_*{C_\phi }} \mathord{\left/
						{\vphantom {{4\pi_*{C_\phi }} {n_i^s}}} \right.
						\kern-\nulldelimiterspace} {n_i^s}}} \right)^2}] \\ &\leqslant \exp ( - {{2{\varepsilon ^2}} \mathord{\left/
				{\vphantom {{2{\varepsilon ^2}} {[{n_i}}}} \right.
				\kern-\nulldelimiterspace} {[{n_i}}}{\left( {{{4\pi _*{C_\phi }} \mathord{\left/
						{\vphantom {{4\pi_*{C_\phi }} {{n_i}}}} \right.
						\kern-\nulldelimiterspace} {{n_i}}}} \right)^2}]
	\end{split}
\end{equation}
Then, for any $\delta > 0 $, with probability at least $1-\delta/2$, the following holds:
\begin{equation}
	\begin{split}
		sup_{\bm f\in \mathcal{H}} &| \widehat R_{i}( {\bm f} ) - R_{i}( {\bm f})| \\& \leqslant \mathbb{E}_{S_{\bm f}} \Phi_{i} (S_{i}) + 2{\pi_*}{C_\phi }\sqrt {\frac{2{\ln (2/\delta )}}{{{n_{i}}}}}
	\end{split}
\end{equation}

By using the Rademacher complexity \cite{DBLP:books/daglib/0034861}, we can obtain 
\begin{equation}
	\begin{split}
		sup_{\bm f\in \mathcal{\bm f}} | \widehat R_{i}( {\bm f} ) - R_{i}( {\bm f})| \leqslant 2 \mathfrak{R}_{{n_{i}}}({ \widetilde l_{i} }{\circ \mathcal H}) \\ + 2{\pi_*}{C_\phi }\sqrt {\frac{2{\ln (2/\delta )}}{{{n_{i}}}}}
	\end{split}
\end{equation} where $\mathfrak{R}_{{n_{i}}}({ \widetilde l_{i} }{\circ \mathcal H})$ is the Rademacher complexity of the composite function class (${ \widetilde l_{i} }{\circ \mathcal H}$) for examples size $n_{i}$. As $L_\phi$ is the Lipschitz constant of $\phi$, we have $\mathfrak{R}_{{n_{i}}}({ \widetilde l_{i} }{\circ \mathcal H}) \leqslant 4{N^s}{L_\phi }{\mathfrak{R}_{{n_{i}}}}(\mathcal{H})$ by Talagrand's contraction Lemma \cite{DBLP:books/daglib/0034861}, where $\widetilde l_{i}(\bar y,\bm f(X)) = \frac{{{n _i}}}{{n_i^s}} [ \phi (  {f_{i}}) - \phi ({f_{C}}) + \phi ( -{f_{C}}) - \phi ({-f_{\bar y}})]$. Then, we can obtain the 
\begin{equation}
	\begin{split}
		{\sup _{\bm f\in \mathcal{H}}}\left| {{R_{i}}(\bm f) -  {\widehat{R}_{i}}  (\bm f)} \right| \leqslant 8{N^s}{L_\phi }{\mathfrak{R}_{{n_{i}}}}(\mathcal{H}) \\ + 2{\pi_*}{C_\phi }\sqrt {\frac{2{\ln (2/\delta )}}{{{n_{i}}}}} 
	\end{split}
\end{equation}

${\sup _{\bm f \in \mathcal{H}}}\left| {{R_u}(\bm f) -  {\widehat{R}_u}  (\bm f)} \right|$ can be proven using the same proof technique, which proves Lemma 5. \hfill $\square$

\section{Proof of Theorem 6}

\textbf{Theorem 6.} For any $\delta>0$, with the probability at least $1-\delta/2$, we have
\begin{equation}
	\begin{split}
		R_{ESA}&({\hat {\bm f}_{ESA})} -  \mathop{\rm {min}} _{{\bm f} \in \mathcal{H}}R_{ESA}(\bm f) \leqslant  \\& \sum\nolimits_{i = 1}^{C - 1} {16{N^s}{L_\phi }{\mathfrak{R}_{{n_{i}}}}(\mathcal{H}) } + 8{N^s}{L_\phi }{C}{\mathfrak{R}_{{n_{u}}}}(\mathcal{H})\\&+  \sum\nolimits_{i = 1}^{C - 1}{4{\pi_{*}}{C_\phi }\sqrt {\frac{{2\ln (2/\delta )}}{{{n_{i}}}}} } + 4{C_\phi }{C}\sqrt {\frac{{2\ln (2/\delta )}}{{{n_{u}}}}} 
	\end{split}
\end{equation} where $\hat {\bm f}_{ESA}$ denotes the trained model by minimizing the ESA risk $\hat{R}_{ESA}  (\bm f)$.

\emph{Proof.} According to Lemma 5, the estimation error bound is proven through 
\begin{align*}
	R_{ESA}&({\widehat {\bm f}_{ESA}}) - R_{ESA}({ \bm f^*})   \\&= ( \widehat R_{ESA}({\widehat {\bm f}_{ESA} }) - \widehat R_{ESA}({\widehat {\bm f^*}})) \\& \; \; \; + (R_{ESA}({\widehat {\bm f}_{ESA}}) - \widehat R_{ESA}({\widehat {\bm f}_{ESA}})) \\& \; \; \; + (\widehat R_{ESA}({\widehat{\bm f^*} }) - R_{ESA}({\widehat {\bm f^*}}))
	\\& \leqslant 0 + 2 sup_{\bm f\in \mathcal{H}} | R_{ESA}( {\bm f}) - \widehat R_{ESA}( {\bm f} ) |
\end{align*} where $ \bm f^* = arg\mathop{\rm {min}}_{{\bm f} \in \mathcal{H}}R(\bm f)$.

We have seen the definition of $ R_{ESA}({ {\bm f}}) $  and $ \widehat R_{ESA}({ {F}}) $ that can also be decomposed into \begin{equation}\label{sre}
	\begin{split}
		R_{ESA}&(f_1,...,f_C) = \sum\limits_{i = 1}^{C - 1} {{\pi _i}{\mathbb{E}_{p^s_i}}\bigg[\phi\big({f_i}(x)\big)}  - \phi\big({f_C}(x)\big) \\& + \big[\phi( - {f_C}(x)) - \phi( - {f_i}(x)\big]\bigg] \\& + {\mathbb{E}_{p^s_u}}\big[\phi({f_C}(x)) + \sum\limits_{y \ne C} {\phi( - {f_y}(x))} \big]
	\end{split}
\end{equation}
and 	
\begin{equation}\label{esre}
	\begin{split}
		\hat R_{ESA}&(f_1,...,f_C) = \sum\limits_{i = 1}^{C - 1} \frac{{{\pi _i}}}{{n_i^s}}\sum\limits_{{x_k} \in {D^s_i}} [\phi({f_i}({x_k})) \\& - \phi( {f_C}({x_k}))  + \phi( - {f_C}({x_k})) - \phi({-f_i}({x_k})]  \\& + \frac{1}{{n_u^s}}\sum\limits_{{x_{k'}} \in {D^s_u}} {[\phi({f_C}({x_{k'}})) + \sum\limits_{y \ne C} {\phi( - {f_y}({x_{k'}}))} ]} 
	\end{split}
\end{equation} Due to the sub-additivity of  the supremum operators with respect to risk, it holds that 
\begin{equation*}
	\begin{split}
		sup_{\bm f\in \mathcal{H}} |\widehat R_{ESA}( {\bm f} )& - R_{ESA}( {\bm f}) | \\ & \leqslant \sum\nolimits_{i = 1}^{C - 1}  sup_{F\in \mathcal{H}} | \widehat R_{i}( {\bm f} ) - R_{i}( {\bm f})|  \\& +  sup_{\bm f\in \mathcal{H}} |\widehat R_{u}( {\bm f} ) - R_u( {\bm f})| 
	\end{split}
\end{equation*}

According to the Lemma 5, we can get the generalization bound that 
\begin{equation}
	\begin{split}
		R_{ESA}&({\hat {\bm f}_{ESA})} -  \mathop{\rm {min}} _{{\bm f} \in \mathcal{H}}R_{ESA}(\bm f) \leqslant  \\& \sum\nolimits_{i = 1}^{C - 1} {16{N^s}{L_\phi }{\mathfrak{R}_{{n_{i}}}}(\mathcal{H}) } + 8{N^s}{L_\phi }{C}{\mathfrak{R}_{{n_{u}}}}(\mathcal{H})\\&+  \sum\nolimits_{i = 1}^{C - 1}{4{\pi_{*}}{C_\phi }\sqrt {\frac{{2\ln (2/\delta )}}{{{n_{i}}}}} } + 4{C_\phi }{C}\sqrt {\frac{{2\ln (2/\delta )}}{{{n_{u}}}}} 
	\end{split}
\end{equation} with probability at least $1-\delta/2$, which finishes the proof. \hfill $\square$

\end{document}